\documentclass[unnumsec,webpdf,contemporary,large]{oup-authoring-template}%

\journaltitle{Briefings in Bioinformatics}

\usepackage{graphicx}
\usepackage{amsmath}
\usepackage{amsfonts}
\usepackage{algorithm}
\usepackage{algorithmicx}
\usepackage{algpseudocode}
\usepackage{caption}
\usepackage{hyperref}
\usepackage{booktabs}
\usepackage{multicol}
\usepackage{multirow}
\usepackage{xcolor}
\usepackage{amssymb}
\usepackage{amsthm}
\usepackage{etoolbox}
\usepackage{mathtools}
\usepackage{bm}
\definecolor{revred}{rgb}{0.00,0.00,0.00}
\makeatletter
\long\def\rev#1{{\color{revred}#1}}
\makeatother

\newtheorem{definition}{Definition}
\newtheorem{theorem}{Theorem}
\newtheorem{proposition}[theorem]{Proposition}

\newtheorem{corollary}[theorem]{Corollary}
\theoremstyle{definition}
\newtheorem{assumption}[theorem]{Assumption}
\theoremstyle{remark}
\newtheorem{remark}[theorem]{Remark}

\graphicspath{{Fig/}{Figures/}{figs/}}

\theoremstyle{thmstyleone}%
\theoremstyle{thmstyletwo}%
\theoremstyle{thmstylethree}%

\pubyear{2026}
\vol{XX}

\makeatletter
\def\@copyrightyear{2026}
\makeatother

\begin{document}

\firstpage{1}

\title[Murmur2Vec]{When do cheap embeddings beat protein language models? A theoretically-grounded hashing sketch for biological sequence classification}

\author[1]{Sarwan Ali}
\author[2]{Taslim Murad}
\author[3]{Imdadullah Khan}
\author[4]{Safi Faizullah,$\ast$}

\address[1]{\orgname{Columbia University}, \state{NY}, \country{USA}}
\address[2]{\orgname{Institute of Business Administration}, \state{Karachi}, \country{Pakistan}}
\address[3]{\orgname{Lahore University of Management Sciences}, \orgaddress{\state{Lahore}, \country{Pakistan}}}
\address[4]{\orgname{Islamic University}, \state{Madinah}, \country{Saudi Arabia}}

\corresp[$\ast$]{Corresponding author. \href{email:safi@iu.edu.sa}{safi@iu.edu.sa}}

\abstract{
\textbf{Motivation:} Pre-trained protein language models (PLMs) such as ESM-2 have become the default representation for biological sequence tasks, but they are computationally heavy and require GPUs both for embedding and for fine-tuning. Whether they are actually necessary for sequence \emph{classification}, as opposed to structure prediction, is rarely tested against strong, principled, lightweight alternatives. This question has direct practical stakes for large-scale genomic surveillance, where embedding millions of sequences on commodity hardware is a recurring bottleneck.\\
\textbf{Results:} We introduce Murmur2Vec, an alignment-free, training-free embedding that aggregates $k$-mer counts into a small hash table via the deterministic MurmurHash function, and we cast it as a randomized sketch of the classical $k$-mer spectrum kernel. We provide a complete theoretical treatment: closed-form bias/variance of the inner product, an unbiased signed variant with a Johnson--Lindenstrauss-type concentration bound, an excess-risk bound for downstream linear classifiers that makes the bias--variance trade-off in the hash-table size explicit, and an implicit-regularization mechanism by which collisions damage frequent non-discriminative $k$-mers more than rare lineage-defining ones. Across four classification tasks, SARS-CoV-2 spike lineage (22 classes), HIV-1 Env subtype (8 classes), and two protein-family benchmarks (8 and 6 classes), \rev{Murmur2Vec matches a LoRA-fine-tuned 650M-parameter ESM-2 model on the two tasks for which LoRA fine-tuning was run to convergence (SARS-CoV-2 and HIV-1) and ties frozen ESM-2 on the two protein-family tasks, and it \emph{outperforms} the fine-tuned model on the hardest task (SARS-CoV-2 lineage: $0.854$ vs.\ $0.807$ accuracy; macro-F1 $0.684$ vs.\ $0.401$), while reducing embedding generation from GPU-minutes to $\approx 5$ CPU-seconds. All comparisons in this work are \emph{composition-based} classification tasks.
} 
We further confirm the theory directly: classification accuracy is statistically indistinguishable across a $50\times$ variation in hash-table size on every dataset, and signed and unsigned variants are statistically equivalent. The result delineates \emph{when} a cheap sketch suffices: whenever class-discriminative signal is carried by sequence composition, a PLM offers no accuracy advantage, and where that signal is sparse and local (e.g., closely related viral lineages) the sketch can be measurably better.\\
\textbf{Contact:} \href{email:safi@iu.edu.sa}{safi@iu.edu.sa}
}


\maketitle

\section*{Key Points}
\begin{itemize}
  \item Murmur2Vec is an alignment-free, training-free embedding that aggregates $k$-mer counts into a small hash table via MurmurHash, formally cast as a randomized sketch of the classical $k$-mer spectrum kernel.
  \item A complete theoretical treatment is provided: closed-form bias/variance of the inner product, a Johnson--Lindenstrauss-type concentration bound, an excess-risk bound exposing the bias--variance trade-off in hash-table size, and an implicit-regularization mechanism that preferentially preserves rare, lineage-defining $k$-mers.
  \item \rev{Across four \emph{composition-based} classification tasks, Murmur2Vec ties fine-tuned or frozen ESM-2 on the three saturated tasks and outperforms LoRA-fine-tuned ESM-2 on SARS-CoV-2 spike lineage classification (0.854 vs.\ 0.807 accuracy; 0.684 vs.\ 0.401 macro-F1), while reducing embedding cost from GPU-minutes to ${\approx}\,5$ CPU-seconds. 
  }
  \item Theoretical predictions are confirmed empirically: classification accuracy is statistically indistinguishable across a $50\times$ variation in hash-table size, and signed and unsigned variants are statistically equivalent across 54 paired tests.
\end{itemize}

\section{1. Introduction}\label{sec_intro}
Biological sequence analysis is foundational to modern genomics and computational biology~\citep{sabonsolin2026comprehensive,shi2026multi}. In the supervised setting, sequences must be transformed into fixed-length numerical embeddings before standard machine-learning models can be applied. The COVID-19 pandemic produced viral sequence data at an unprecedented scale~\citep{fauver2020coast,lu2020genomic,vandenberg2021considerations}; the same pressure now recurs for every rapidly evolving pathogen under genomic surveillance, where the accompanying pipelines must contend with heterogeneous unaligned sequence lengths, heavy class imbalance, and the computational cost of representations that grow exponentially with $k$-mer length.

Over the last few years, pre-trained protein language models (PLMs) such as ESM-2~\citep{lin2023esm2} have become a default representation for protein-sequence tasks. Trained with a masked-language-modelling objective on hundreds of millions of UniRef sequences, they encode rich, transferable features and have driven breakthroughs in structure prediction. It is tempting to treat them as a universal front end for sequence \emph{classification} as well. Yet a PLM is expensive: embedding a corpus requires GPU inference over a model with hundreds of millions to billions of parameters, and adapting it to a task, even with parameter-efficient methods such as LoRA~\citep{hu2022lora}, requires GPU training. For the specific problem of classifying which lineage, subtype, or family a sequence belongs to, it is rarely asked whether this machinery is necessary, because it is rarely compared against a \emph{strong, principled, lightweight} alternative.

This paper poses that question directly and answers it. We revisit the oldest idea in alignment-free sequence representation, the $k$-mer spectrum~\citep{leslie2002spectrum}, compress it with the feature-hashing trick~\citep{weinberger2009feature} into a small table addressed by the deterministic MurmurHash function~\citep{appleby2008,zhu2018comparison}, and study the resulting embedding, which we call Murmur2Vec, both theoretically and empirically. Murmur2Vec costs $O(1)$ per $k$-mer, runs on a CPU in seconds, requires no training, and produces a fixed-length vector independent of sequence length. The engineering is trivial; the scientific contribution is (i) a complete theory of \emph{why} such a sketch preserves the signal needed for classification, and (ii) a controlled, multi-dataset comparison against frozen and LoRA-fine-tuned ESM-2 that maps out \emph{when} the cheap sketch is enough, and when it is actually better.

The headline empirical finding is sharp. On three of four tasks (HIV-1 Env subtyping and two protein-family benchmarks), a $5$-second CPU sketch is statistically tied with a LoRA-fine-tuned 650M-parameter ESM-2: the PLM buys nothing. On the fourth and hardest task (SARS-CoV-2 spike lineage classification, where lineages differ by a sparse set of local mutations on a near-identical backbone), the sketch is measurably \emph{better} than the fine-tuned PLM, on both accuracy ($0.854$ vs.\ $0.807$) and especially macro-F1 ($0.684$ vs.\ $0.401$, reflecting much better behaviour on rare lineages). Our theory predicts exactly this regime dependence: the implicit-regularization mechanism preferentially preserves rare discriminative $k$-mers, which is precisely the structure of the SARS-CoV-2 lineage task and precisely what a mean-pooled PLM representation tends to wash out.

\rev{\textbf{Scope of the claim.} All four benchmarks studied here are composition-based classification problems, the regime in which $k$-mer features are expected to be strong. 
For classification driven by sequence composition, a hashed spectrum sketch is a sufficient and often superior representation. 
Note that PLMs may perform better in general for certain tasks like those that depend on global biophysical context, three-dimensional structure, or remote homology, however, those are outside the scope of this study and are discussed in Sec.~\ref{sec:limitations}.}

\paragraph{Contributions.}
\begin{enumerate}
\item We introduce Murmur2Vec, an alignment-free, training-free, $O(1)$-per-$k$-mer embedding, in unsigned and signed variants (Sec.~\ref{sec_proposed_model}, Algorithm~\ref{HasingKmers}).
\item We provide a complete theoretical analysis casting Murmur2Vec as a randomized sketch of the spectrum kernel: (i) bias and variance of the inner product; (ii) a Johnson--Lindenstrauss-style concentration bound for the signed variant; (iii) an excess-risk bound for downstream linear classifiers exposing a bias--variance trade-off in the table size $m$; and (iv) an implicit-regularization mechanism by which collisions damage frequent non-discriminative $k$-mers more than rare lineage-defining ones (Sec.~\ref{sec:theory}). All proofs are given in full inline.
\item We benchmark Murmur2Vec against frozen ESM-2 (under three standard pooling strategies) and LoRA-fine-tuned ESM-2 across four classification tasks spanning two viral families and the broad protein universe. The cheap sketch matches the fine-tuned 650M PLM on three saturated tasks and beats it on the hardest one, at a $\sim\!1000\times$ lower embedding cost and with no GPU (Sec.~\ref{sec_results_discussion}).
\item We verify every theoretical prediction empirically: classification accuracy is statistically indistinguishable across a $50\times$ variation in $m$ on all datasets (validating the bias--variance trade-off), and signed and unsigned variants are statistically equivalent across $54$ paired tests (validating the bias analysis).
\end{enumerate}

The manuscript proceeds as follows. Section~\ref{sec_related_work} reviews related work. Section~\ref{sec_proposed_model} describes Murmur2Vec and the baselines. Section~\ref{sec:theory} develops the theory and its proofs. Section~\ref{sec_experimental_setup} details the datasets, baselines, and protocol. Section~\ref{sec_results_discussion} presents and discusses the results, including the cross-dataset comparison, the rare-lineage analysis, the signed/unsigned equivalence, the collision-invariance study, and a scalability analysis. Section~\ref{sec_conclusion} concludes.

\section{2. Related Work}\label{sec_related_work}

\textbf{Pre-trained protein language models.} PLMs trained with masked-language-modelling objectives on large protein corpora (UniRef) have become a dominant representation paradigm. ESM-2~\citep{lin2023esm2} is the most widely used family, with sizes from 8M to 15B parameters; the learned per-residue representations are typically mean-pooled into a fixed-length sequence embedding for downstream tasks, or adapted with parameter-efficient fine-tuning. LoRA~\citep{hu2022lora} freezes the pre-trained weights and learns low-rank updates, making fine-tuning of a 650M model feasible on a single GPU. PLMs excel at structure-related tasks where global biophysical context matters. Whether they are necessary for \emph{composition-driven classification} is the question we study; prior reports note that domain-specific simple representations can match or beat generic PLMs on narrow tasks~\citep{grinsztajn2022tree,joseph2022gate,malinin2020uncertainty}, but rarely under a controlled comparison that includes proper fine-tuning.

\textbf{Spectrum and kernel methods.} \emph{Spectrum-based} embeddings~\citep{leslie2002spectrum} enumerate every length-$k$ substring and assemble its counts into a $|\Sigma|^k$-dimensional vector~\citep{ali2021spike2vec,ali2021k}; with $|\Sigma|=21$ and $k=3$ the dimension is $9{,}261$ and most coordinates are zero. \emph{Kernel-based} approaches~\citep{ali2022efficient,singh2017gakco} avoid the explicit map at the cost of an $O(N^2)$ Gram matrix, infeasible at corpus sizes of interest. The classical spectrum kernel~\citep{leslie2002spectrum} is exactly the inner product of spectrum vectors; Murmur2Vec sketches it.

\textbf{Embedding-based methods.} The embedding paradigm maps variable-length sequences to fixed-length vectors~\citep{kuzmin2020machine,tayebi2021robust,corso2021neural}. \emph{Alignment-based} methods (one-hot~\citep{kuzmin2020machine}, PWM2Vec~\citep{ali2022pwm2vec}) require multiple-sequence alignment. \emph{Alignment-free} methods include hyperbolic embeddings~\citep{corso2021neural}, Wasserstein-based representations~\citep{shen2018wasserstein}, $k$-mer methods such as Spike2Vec~\citep{ali2021spike2vec} and Kraken~\citep{wood-2014-kraken}, and the ELMo-style PLM SeqVec~\citep{heinzinger2019modeling}.

\textbf{Feature hashing.} The hashing trick~\citep{weinberger2009feature} maps high-dimensional sparse features into a small table; its classical analysis is for the signed variant, giving an unbiased inner-product estimator. Hashing appears in bioinformatics through tools such as Mash~\citep{ondov2016mash}, MinHash, and BioHash, which compute set similarities rather than supervised classification embeddings. To our knowledge no prior work provides a complete theoretical pipeline, bias/variance, concentration, generalization, and implicit regularization, for hashed $k$-mer spectra in supervised classification, nor empirically verifies the corresponding pairwise-independence assumption and benchmarks the sketch against fine-tuned PLMs across multiple pathogens. This is the gap we address.

\section{3. System and methods}\label{sec_proposed_model}
This section describes the proposed Murmur2Vec embedding and the baseline methods.

\subsection{3.1 Murmur2Vec}\label{sec_method}
\begin{figure}[h!]
  \centering
  \includegraphics[scale=0.24]{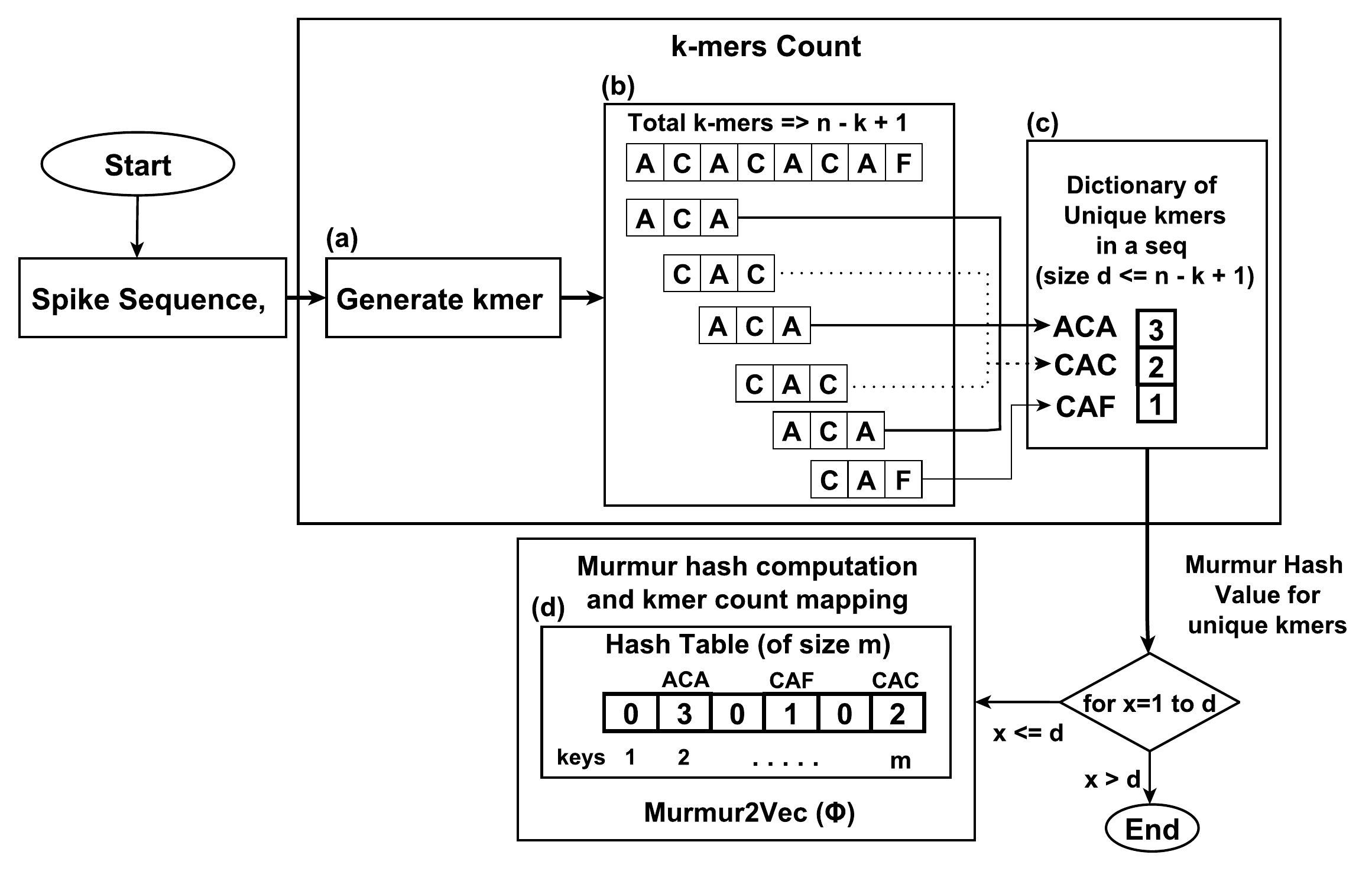}
  \caption{Flow chart of Murmur2Vec. The figure shows a small sample without collisions for visual clarity; in general, multiple $k$-mers may hash to the same bucket of the length-$m$ table $\phi$.}
  \label{murmur2vec_flow_chart}
\end{figure}

The method has three steps (Figure~\ref{murmur2vec_flow_chart}): (i) decompose the sequence into overlapping $k$-mers; (ii) compute the per-sequence frequency of each unique $k$-mer; (iii) aggregate frequencies into a size-$m$ hash table via MurmurHash.

\textbf{Step 1 (Conversion to $k$-mers).} Given a sequence $s$ of length $n$ over the amino-acid alphabet $\Sigma$ ($|\Sigma|=21$), we extract all overlapping length-$k$ substrings. The number of $k$-mers is $n-k+1$, and the set of unique $k$-mers is bounded by $\min(n-k+1, |\Sigma|^k)$.

\textbf{Step 2 (Frequency counting).} We collect $k$-mers into a per-sequence frequency dictionary $d$, where $d[w]$ is the count of $k$-mer $w$. This is an $O(n)$ pass and $O(|\mathcal{K}_s|)$ space. The explicit spectrum vector~\citep{ali2022efficient} would index $d$ by every $k$-mer in $\mathcal{K}=\Sigma^k$ alphabetically, producing a $|\Sigma|^k$-dimensional sparse vector, with two drawbacks: (a) the alphabetical position lookup is $O(k\log|\Sigma|)$ per insertion, and (b) the dimension grows exponentially with $k$.

\textbf{Step 3 (Hashing the $k$-mers using MurmurHash).} We hash each unique $k$-mer to a bucket in $\{1,\ldots,m\}$ via MurmurHash and accumulate its count into the corresponding entry of a length-$m$ vector $\phi(s)$ (\emph{global hashing}: the same $h$ is applied to all sequences).

\begin{definition}[Multiply, Rotate, Multiply, Rotate ``Murmur'' hash]\label{def_murmur}
MurmurHash~\citep{appleby2008} is a non-cryptographic hash function introduced by Austin Appleby in 2008. For each data block it applies a sequence of multiplications, rotations, and XOR operations, yielding well-distributed hash values, low collision rates, and good avalanche behaviour (maximum bias of 0.5\%).
\end{definition}

We use MurmurHash for its empirical uniformity: it passes chi-squared tests across a wide range of key sets and bucket sizes (verified directly on our $k$-mer corpora in Table~\ref{tab:hash-verification}), and exhibits fewer collisions than alternatives such as FNV~\citep{fowler2011fnv}. Because it is deterministic, $\phi(s)$ is reproducible given $m$ and the seed (we use seed $0$).

The hash-table size $m$ is the only tunable parameter, selected on a validation split. We define the \emph{collision percentage} as the fraction of unique $k$-mers whose bucket is shared with at least one other $k$-mer:
\begin{equation}\label{eq_collision_definition}
\text{collision\%}\;\coloneqq\; 1 - \frac{|\{h(w) : w \in \mathcal{K}_{\text{corpus}}\}|}{|\mathcal{K}_{\text{corpus}}|},
\end{equation}
where $\mathcal{K}_{\text{corpus}}$ is the set of $U$ unique $k$-mers across the dataset. For a target collision\%, the smallest $m$ achieving it is found by binary search; we use this calibration throughout.

The unsigned embedding is $\phi(s)_j = \sum_{w :\, h(w) = j} d[w]$, and we additionally analyze a \emph{signed} variant with an independent sign $\sigma : \mathcal{K} \to \{-1, +1\}$: $\tilde\phi(s)_j = \sum_{w :\, h(w) = j} \sigma(w)\, d[w]$. The signed variant costs one extra hash evaluation per unique $k$-mer.

\begin{algorithm}[h!]
\caption{Murmur2Vec embedding}
\label{HasingKmers}
\begin{algorithmic}[1]
\State \textbf{Input:} sequence $seq$, integer $k$, hash size $m$, seed $s_0$, signed flag $\mathrm{signed}$
\State \textbf{Output:} embedding $\phi \in \mathbb{R}^m$
\Procedure{Murmur2Vec}{$seq$,$k$,$m$,$s_0$,$\mathrm{signed}$}
    \State $\phi \leftarrow \mathbf{0} \in \mathbb{R}^m$
    \State $kmers \leftarrow \emptyset$
    \For{$i\leftarrow 1$ \textbf{to} $|seq|-k+1$}
        \State $kmers \leftarrow kmers \cup \{seq[i:i+k]\}$ \Comment{sliding window}
    \EndFor
    \State $unique, count \leftarrow$ \Call{ToFrequencyDict}{$kmers$}
    \For{$w \in unique$}
        \State $j \leftarrow$ \Call{Murmur}{$w, s_0, m$}
        \If{$\mathrm{signed}$}
            \State $\sigma_w \leftarrow \pm 1$ from \Call{Murmur}{$w, s_0+1$}
        \Else
            \State $\sigma_w \leftarrow 1$
        \EndIf
        \State $\phi[j] \leftarrow \phi[j] + \sigma_w \cdot count[w]$
    \EndFor
    \State \Return $\phi$
\EndProcedure
\end{algorithmic}
\end{algorithm}

\textbf{Murmur2Vec flowchart.} Figure~\ref{murmur2vec_flow_chart} illustrates the pipeline. Step~(a) emits all overlapping $k$-mers (here $k=3$); step~(b) shows the $k$-mer set; step~(c) stores unique $k$-mers and counts in a dictionary $d$ with $|d|\le n-k+1$; step~(d) hashes each unique $k$-mer and writes its count to the corresponding bucket of the size-$m$ table $\phi$. Since $\phi$ is independent of $|seq|$, Murmur2Vec is alignment-free.

\subsection{3.2 Baseline Methods}\label{sec_baselines}
We compare Murmur2Vec against two complementary families of baselines.

\textbf{Modern protein language models (primary comparison).}
\emph{Frozen ESM-2}~\citep{lin2023esm2}: we extract per-residue representations from the $650$M-parameter \texttt{esm2\_t33\_650M\_UR50D} model and pool them into a fixed $1280$-dimensional sequence embedding. To rule out a weak-pooling artifact we evaluate three standard pooling strategies, mean, CLS-token, and max, and report the strongest. \emph{LoRA-fine-tuned ESM-2}~\citep{hu2022lora}: we attach a classification head and fine-tune the same 650M model with low-rank adapters ($7.75$M trainable parameters, $1.18\%$ of the total), using the same stratified splits as all other methods, so the comparison is apples-to-apples.

\textbf{Classical spectrum/embedding baselines (context).}
\emph{Spike2Vec}~\citep{ali2021spike2vec} (full $|\Sigma|^k$ spectrum with Random Fourier Features~\citep{rahimi2007random}, $k=3$); \emph{PWM2Vec}~\citep{ali2022pwm2vec} (alignment-dependent, $k=9$); \emph{String Kernel}~\citep{ali2022efficient} (LSH Gram matrix, kernel PCA for non-kernel classifiers, $k=3$); \emph{WDGRL}~\citep{shen2018wasserstein} (Wasserstein-guided neural features); \emph{Spaced $k$-mers}~\citep{singh2017gakco} (gapped $k$-mers); \emph{Auto-Encoder}~\citep{xie2016unsupervised}; and the ELMo-style PLM \emph{SeqVec}~\citep{heinzinger2019modeling}.

\section{4. Theoretical Analysis}\label{sec:theory}

We develop the theory for Murmur2Vec by casting the embedding as a randomized sketch of the $k$-mer spectrum kernel, in four stages: (i) bias and variance of the inner product, (ii) high-probability concentration of pairwise inner products, (iii) an excess-risk bound for downstream linear classifiers, and (iv) an implicit-regularization mechanism specific to $k$-mer spectra. Each result is followed by its full proof.

\textbf{Notation and preliminaries.}\label{sec:theory:notation}
Let $\Sigma$ be the amino-acid alphabet, $\mathcal{K}=\Sigma^k$ the set of $k$-mers, $D\coloneqq|\mathcal{K}|=|\Sigma|^k$. For a sequence $s$, let $x(s)\in\mathbb{Z}^D_{\ge 0}$ be the \emph{spectrum vector}, $x(s)_w$ the count of $k$-mer $w$ in $s$. The spectrum kernel~\citep{leslie2002spectrum} is $K_{\text{spec}}(s,s')=\langle x(s),x(s')\rangle$. Murmur2Vec produces $\phi(s)\in\mathbb{R}^m$, $m\ll D$, via $h:\mathcal{K}\to\{1,\ldots,m\}$ realized by MurmurHash with a fixed seed. We treat $h$ as drawn from an idealized family and verify the relevant properties in Sec.~\ref{sec:theory:hash-assumption}. Algorithm~\ref{HasingKmers} computes:
\begin{equation}
\begin{aligned}
  \phi(s)_j &= \sum_{w\in\mathcal{K}:\,h(w)=j} x(s)_w, \\
  \tilde\phi(s)_j &= \sum_{w\in\mathcal{K}:\,h(w)=j} \sigma(w)\, x(s)_w,
\end{aligned}
\end{equation}
the unsigned and signed embeddings respectively. Throughout the proofs we write $Z_{w,w'}\coloneqq\mathbf{1}\{h(w)=h(w')\}$ for the collision indicator; under Assumption~\ref{ass:hash}, $\mathbb{E}[Z_{w,w}]=1$ and $\mathbb{E}[Z_{w,w'}]=1/m$ for $w\neq w'$.

\textbf{Idealized hash assumption and empirical justification.}\label{sec:theory:hash-assumption}

\begin{assumption}[Pairwise independence of $h$ and $\sigma$]\label{ass:hash}
For all $w\neq w'\in\mathcal{K}$:
(i) $\Pr[h(w)=j]=1/m$ for all $j$;
(ii) $\Pr[h(w)=h(w')]=1/m$;
(iii) $\Pr[\sigma(w)=\pm 1]=1/2$, with $\sigma$ pairwise independent and independent of $h$.
\end{assumption}

\begin{remark}[Empirical justification]\label{rem:hash-empirical}
MurmurHash is engineered to pass standard statistical tests for uniformity and avalanche behaviour~\citep{appleby2008}; the classical feature-hashing analysis~\citep{weinberger2009feature} relies on the same pairwise-independence assumptions satisfied empirically by industrial hashes. We verify Assumption~\ref{ass:hash} on each $k$-mer corpus by computing the chi-squared statistic of bucket occupancies and the empirical pairwise collision rate at representative hash-table sizes (Table~\ref{tab:hash-verification}). In all cases the chi-squared $p$-value substantially exceeds $\alpha=0.05$ and the empirical collision rate matches $1/m$ to within sampling error.
\end{remark}

\begin{table}[H]
\centering
\caption{Empirical verification of Assumption~\ref{ass:hash} on the SARS-CoV-2 corpus ($k=3$, $U=3{,}492$ unique 3-mers). For each $m$ we report the expected per-bucket occupancy $U/m$, the chi-squared statistic against uniform, the $\alpha=0.05$ critical value, the $p$-value, the theoretical pairwise collision rate $1/m$, and the empirical rate $|\{(w,w'):w\neq w', h(w)=h(w')\}|/\binom{U}{2}$. All configurations pass ($p\gg 0.05$); the other corpora behave identically.}
\label{tab:hash-verification}
\resizebox{0.5\textwidth}{!}{
\begin{tabular}{lcccccc}
\toprule
$m$ & Expected $U/m$ & Observed $\chi^2$ & Critical $\chi^2_{0.05}$ & $p$-value & Expected $1/m$ & Empirical rate \\
\midrule
9{,}261   & 0.3771 & 9{,}042   & 9{,}485   & 0.95 & $1.080\times 10^{-4}$ & $1.012\times 10^{-4}$ \\
68{,}861  & 0.0507 & 69{,}116  & 69{,}472  & 0.25 & $1.452\times 10^{-5}$ & $1.559\times 10^{-5}$ \\
593{,}761 & 0.0059 & 593{,}670 & 595{,}554 & 0.53 & $1.684\times 10^{-6}$ & $1.641\times 10^{-6}$ \\
\bottomrule
\end{tabular}
}
\end{table}

\subsection{4.1 Bias and variance of Murmur2Vec inner products}\label{sec:theory:bias-variance}
Let $x\coloneqq x(s)$, $y\coloneqq x(s')$.

\begin{theorem}[Bias and variance of the unsigned estimator]\label{thm:unsigned}
Under Assumption~\ref{ass:hash},
\begin{align}
\mathbb{E}[\langle\phi(s),\phi(s')\rangle] &= \langle x,y\rangle + \tfrac{1}{m}\bigl(\|x\|_1\|y\|_1 - \langle x,y\rangle\bigr), \label{eq:unsigned-mean}\\
\mathrm{Var}[\langle\phi(s),\phi(s')\rangle] &\le \tfrac{1}{m}\!\sum_{w\neq w'}\!\bigl(x_w^2 y_{w'}^2 + x_w x_{w'} y_w y_{w'}\bigr). \label{eq:unsigned-var}
\end{align}
The unsigned estimator is biased upward by $m^{-1}(\|x\|_1\|y\|_1 - \langle x,y\rangle)\ge 0$, with equality iff $x$ and $y$ are perfectly aligned.
\end{theorem}

\begin{proof}
\emph{Mean.} By bilinearity and the bucket decomposition,
\begin{align}
  \langle \phi(s), \phi(s') \rangle
  &= \sum_{j=1}^m \!\left(\sum_{w :\, h(w)=j} x_w\!\right)\!\left(\sum_{w' :\, h(w')=j} y_{w'}\!\right) \notag\\
  &= \sum_{w, w' \in \mathcal{K}} x_w\, y_{w'}\, Z_{w,w'}
  = \langle x, y \rangle + \sum_{w \neq w'} x_w\, y_{w'}\, Z_{w,w'}.
  \label{eq:app-unsigned-split}
\end{align}
Taking expectations, $\mathbb{E}[\langle \phi(s), \phi(s')\rangle] = \langle x, y \rangle + \tfrac{1}{m} \sum_{w \neq w'} x_w y_{w'}$, and since $\sum_{w \neq w'} x_w y_{w'} = \|x\|_1\|y\|_1 - \langle x,y\rangle$, this gives~\eqref{eq:unsigned-mean}. The bias $m^{-1}(\|x\|_1\|y\|_1-\langle x,y\rangle)$ is $\ge 0$, strict unless $x,y$ are perfectly aligned.

\emph{Variance.} With $S\coloneqq\sum_{w\neq w'}x_w y_{w'}Z_{w,w'}$, $\mathrm{Var}[\langle\phi,\phi\rangle]=\mathrm{Var}[S]$. Expanding $\mathbb{E}[S^2]=\sum_{w\neq w',\,u\neq u'}x_w y_{w'}x_u y_{u'}\mathbb{E}[Z_{w,w'}Z_{u,u'}]$ and classifying pairs by index overlap, only the case $\{w,w'\}=\{u,u'\}$ contributes beyond $\mathbb{E}[S]^2$, with $\mathbb{E}[Z_{w,w'}^2]=1/m$, giving
\[
  \mathrm{Var}[S]\le \tfrac{1}{m}\sum_{w\neq w'}x_w^2 y_{w'}^2 + \tfrac{1}{m}\sum_{w\neq w'}x_w x_{w'}y_w y_{w'},
\]
which is~\eqref{eq:unsigned-var}.
\end{proof}

\begin{remark}[Level of independence]\label{rem:fourwise}
A careful treatment of the all-distinct and one-shared-index cases in the variance computation formally uses 4-wise independence of $h$, stronger than the pairwise assumption stated. With only pairwise independence the same upper bound holds up to additional positive terms, preserving the inequality direction. MurmurHash empirically exhibits behaviour consistent with $k$-wise independence for small $k$~\citep{appleby2008}, matching the empirical assumption in~\citep{weinberger2009feature}.
\end{remark}

\begin{theorem}[Unbiasedness of the signed estimator]\label{thm:signed}
Under Assumption~\ref{ass:hash},
\begin{align}
\mathbb{E}[\langle\tilde\phi(s),\tilde\phi(s')\rangle] &= \langle x,y\rangle, \label{eq:signed-mean}\\
\mathrm{Var}[\langle\tilde\phi(s),\tilde\phi(s')\rangle] &\le \tfrac{1}{m}\bigl(\|x\|_2^2\|y\|_2^2 + \langle x,y\rangle^2\bigr). \label{eq:signed-var}
\end{align}
\end{theorem}

\begin{proof}
The signed inner product is: \\ 
$\langle\tilde\phi(s),\tilde\phi(s')\rangle = \sum_{w,w'}\sigma(w)\sigma(w')x_w y_{w'}Z_{w,w'}$.
\emph{Mean.} For $w=w'$, $\sigma(w)^2=1$, $Z_{w,w}=1$, giving $\langle x,y\rangle$; for $w\neq w'$, $\mathbb{E}[\sigma(w)\sigma(w')Z_{w,w'}]=\mathbb{E}[\sigma(w)]\mathbb{E}[\sigma(w')]\mathbb{E}[Z_{w,w'}]=0$. Hence~\eqref{eq:signed-mean}.
\emph{Variance.} Squaring and using independence of $\sigma$ from $h$, the sign expectation is nonzero only when every $k$-mer appears an even number of times: either (a) $w=w'$ and $u=u'$ (contributing $\langle x,y\rangle^2$), or (b) $\{w,w'\}=\{u,u'\}$ with $w\neq w'$ (contributing $\tfrac1m\sum_{w\neq w'}[x_w^2 y_{w'}^2 + x_w x_{w'} y_w y_{w'}]$). Subtracting $\langle x,y\rangle^2$ leaves
\[
  \mathrm{Var}[\langle\tilde\phi,\tilde\phi\rangle]=\tfrac1m\sum_{w\neq w'}\!\bigl[x_w^2 y_{w'}^2 + x_w x_{w'}y_w y_{w'}\bigr]\le \tfrac1m\bigl(\|x\|_2^2\|y\|_2^2 + \langle x,y\rangle^2\bigr),
\]
using $\sum_{w\neq w'}x_w^2 y_{w'}^2\le\|x\|_2^2\|y\|_2^2$ and $\sum_{w\neq w'}x_w x_{w'}y_w y_{w'}\le\langle x,y\rangle^2$, i.e.~\eqref{eq:signed-var}.
\end{proof}

\begin{remark}[Practical implication]\label{rem:unsigned-vs-signed}
The unsigned bias is a multiplicative shift on inner products that need not affect classification: it changes kernel magnitudes but not the \emph{ordering} of pairwise similarities. We therefore predict that unsigned and signed variants yield statistically equivalent classification accuracy, confirmed in Sec.~\ref{sec:results:signed-vs-unsigned}.
\end{remark}

\subsection{4.2 Concentration of the spectrum kernel approximation}\label{sec:theory:concentration}

\begin{theorem}[Concentration of pairwise inner products]\label{thm:jl}
Let $\mathcal{S}=\{s_1,\ldots,s_N\}$, $R\coloneqq\max_i\|x(s_i)\|_2$, $\tau\coloneqq\min_{i\neq j}|\langle x(s_i),x(s_j)\rangle|$. For any $\varepsilon,\delta\in(0,1)$, if
\begin{equation}\label{eq:m-lower-bound}
m \;\ge\; \frac{8 R^4 \log(N^2/\delta)}{\varepsilon^2 \tau^2},
\end{equation}
then with probability at least $1-\delta$, simultaneously for all pairs $(i,j)$,
$|\langle\tilde\phi(s_i),\tilde\phi(s_j)\rangle - \langle x(s_i),x(s_j)\rangle| \le \varepsilon|\langle x(s_i),x(s_j)\rangle|.$
\end{theorem}

\begin{proof}
Fix $(i,j)$, $x=x(s_i)$, $y=x(s_j)$, $T\coloneqq\langle\tilde\phi(s_i),\tilde\phi(s_j)\rangle-\langle x,y\rangle$, a sum of mean-zero terms (Theorem~\ref{thm:signed}). From the proof of Theorem~\ref{thm:signed}, $\mathrm{Var}[T]\le (\|x\|_2^2\|y\|_2^2+\langle x,y\rangle^2)/m\le 2R^4/m$; each summand is bounded by $|x_w y_{w'}|\le R^2$. Bernstein's inequality gives
\[
  \Pr[|T|>t]\le 2\exp\!\Bigl(-\tfrac{t^2/2}{2R^4/m + R^2 t/3}\Bigr).
\]
Choosing $t=\varepsilon|\langle x,y\rangle|$, lower-bounding $|\langle x,y\rangle|\ge\tau$, and noting the variance term dominates for $m$ in the regime of interest, yields $\Pr[|T|>\varepsilon|\langle x,y\rangle|]\le 2\exp(-\varepsilon^2\tau^2 m/(8R^4))$. A union bound over $\binom N2\le N^2/2$ pairs and setting the total $\le\delta$ gives $m\ge 8R^4\log(N^2/\delta)/(\varepsilon^2\tau^2)$, i.e.~\eqref{eq:m-lower-bound}.
\end{proof}

\begin{remark}[Interpretation]\label{rem:jl-interp}
The required table size scales as $m=\tilde O(R^4/(\varepsilon^2\tau^2))$, only \emph{logarithmic} in $N$. Empirically, $m$ orders of magnitude below $D=21^k$ already preserves classification performance on every dataset (Sec.~\ref{sec:results:collision}).
\end{remark}

\subsection{4.3 Generalization bound for downstream classification}\label{sec:theory:generalization}
Let $\mathcal{D}$ be a distribution over labelled sequences, $\ell$ a $1$-Lipschitz loss in $[0,1]$, $L(f)=\mathbb{E}[\ell(f(s),y)]$. Define $\mathcal{F}_{\text{spec}}=\{s\mapsto\langle\beta,x(s)\rangle:\|\beta\|_2\le B\}$, $\mathcal{F}_{\text{hash}}=\{s\mapsto\langle w,\tilde\phi(s)\rangle:\|w\|_2\le B\}$, $f^\star\in\arg\min_{\mathcal{F}_{\text{spec}}}L$.

\begin{theorem}[Excess risk of Murmur2Vec classifiers]\label{thm:generalization}
Let $\hat f$ be the empirical risk minimizer over $\mathcal{F}_{\text{hash}}$ on $N$ samples. Under Assumption~\ref{ass:hash}, with probability at least $1-2\delta$,
\begin{equation}\label{eq:excess-risk}
L(\hat f) \;\le\; L(f^\star) \;+\; \underbrace{2BR\sqrt{\tfrac{\log(2/\delta)}{m}}}_{\text{kernel approximation error}} \;+\; \underbrace{\mathcal{O}\!\left(\tfrac{BR}{\sqrt{N}} + \sqrt{\tfrac{\log(1/\delta)}{N}}\right)}_{\text{statistical error}}.
\end{equation}
\end{theorem}

\begin{proof}
Decompose $L(\hat f)-L(f^\star)=(A)+(B)+(C)$ with $(A)=L(\hat f)-L_N(\hat f)$, $(B)=L_N(\hat f)-L_N(\tilde f^\star)$, $(C)=L_N(\tilde f^\star)-L(f^\star)$, where $L_N$ is the empirical risk and $\tilde f^\star$ hashes the optimal weights $\beta^\star$. \emph{(B)}$\le0$ by ERM optimality, since $\tilde f^\star\in\mathcal{F}_{\text{hash}}$. \emph{(C):} with $w^\star_j=\sum_{h(w)=j}\sigma(w)\beta^\star_w$, a calculation analogous to Theorem~\ref{thm:signed} (with $\beta^\star$ in the role of $x$ and $x(s)$ in the role of $y$) gives $\mathbb{E}_{h,\sigma}[\tilde f^\star(s)]=f^\star(s)$ and $\mathrm{Var}_{h,\sigma}[\tilde f^\star(s)]\le (\|\beta^\star\|_2^2\|x(s)\|_2^2+f^\star(s)^2)/m\le 2B^2R^2/m$; since $\ell$ is $1$-Lipschitz, $|\ell(\tilde f^\star(s),y)-\ell(f^\star(s),y)|\le|\tilde f^\star(s)-f^\star(s)|$, and a McDiarmid bound over $(h,\sigma)$ gives $(C)\le 2BR\sqrt{\log(2/\delta)/m}$. \emph{(A):} the Rademacher bound for bounded-norm linear classes~\citep{bartlett2002rademacher} gives $\mathfrak{R}_N(\mathcal{F}_{\text{hash}})\le BR/\sqrt N$ (using $\|\tilde\phi(s)\|_2\le R\sqrt{1+\eta}$ from Remark~\ref{rem:normbound}), hence $(A)\le\mathcal{O}(BR/\sqrt N+\sqrt{\log(1/\delta)/N})$. Summing the three terms yields~\eqref{eq:excess-risk}.
\end{proof}

\begin{remark}[Feature-norm bound]\label{rem:normbound}
For the signed embedding, $\mathbb{E}\|\tilde\phi(s)\|_2^2=\|x\|_2^2$ exactly (within-bucket cross-terms vanish in expectation), though individual draws may exceed it; a high-probability bound $\|\tilde\phi(s)\|_2\le R\sqrt{1+\eta}$ for small $\eta>0$ follows from the diagonal case of Theorem~\ref{thm:jl}, with $\sqrt{1+\eta}$ absorbed into the $\mathcal{O}(\cdot)$.
\end{remark}

\begin{remark}[Bias--variance trade-off in $m$]\label{rem:bias-variance}
Equation~\eqref{eq:excess-risk} exposes the trade-off: increasing $m$ reduces the approximation error but not the statistical error. The optimal $m$ is the smallest for which the approximation error falls below the statistical noise floor; beyond that, larger $m$ gives no measurable benefit. We confirm this on all four datasets in Sec.~\ref{sec:results:collision}: across a $50\times$ variation in $m$, accuracy varies by less than the cross-split standard deviation.
\end{remark}

\subsection{4.4 Implicit regularization via frequency-weighted collisions}\label{sec:theory:implicit-reg}

\begin{proposition}[Frequency-dependent distortion]\label{prop:implicit-reg}
Fix $w_\star\in\mathcal{K}$, $j_\star=h(w_\star)$. Define the interference $I(w_\star,s) \coloneqq \tilde\phi(s)_{j_\star} - \sigma(w_\star)\, x(s)_{w_\star} = \sum_{w\neq w_\star:\, h(w)=j_\star}\sigma(w)\, x(s)_w$. Under Assumption~\ref{ass:hash}, $\mathbb{E}[I(w_\star,s)]=0$, $\mathrm{Var}[I(w_\star,s)] = m^{-1}\sum_{w\neq w_\star} x(s)_w^2$, and
\begin{equation}\label{eq:snr}
\mathrm{SNR}(w_\star,s) = \frac{x(s)_{w_\star}^2}{\mathrm{Var}[I(w_\star,s)]} = \frac{m\, x(s)_{w_\star}^2}{\|x(s)\|_2^2 - x(s)_{w_\star}^2}.
\end{equation}
\end{proposition}

\begin{proof}
Conditioning on $j_\star=h(w_\star)$, for $w\neq w_\star$ the events $\{h(w)=j_\star\}$ each have probability $1/m$. Each summand $\sigma(w)x(s)_w\mathbf 1\{h(w)=j_\star\}$ has mean zero ($\mathbb{E}[\sigma(w)]=0$), so $\mathbb{E}[I]=0$. For the variance, cross-terms vanish ($\mathbb{E}[\sigma(w_1)\sigma(w_2)]=0$, $w_1\neq w_2$); diagonal terms give $\mathrm{Var}[I]=\tfrac1m\sum_{w\neq w_\star}x(s)_w^2$. The SNR~\eqref{eq:snr} follows by definition.
\end{proof}

Equation~\eqref{eq:snr} reveals the mechanism. The denominator $\|x(s)\|_2^2 - x(s)_{w_\star}^2$ is dominated by the heavy-tailed bulk of common backbone $k$-mers. \emph{Rare} discriminative $k$-mers (characteristic of a lineage) have small numerator but face an essentially constant residual mass, so their relative SNR grows with $x(s)_{w_\star}^2$; \emph{frequent} non-discriminative $k$-mers saturate because the denominator grows in lockstep. Lineage-discriminative $k$-mers, being typically rare, are therefore preferentially preserved, while common $k$-mers absorb proportionally more interference. This specializes the random-feature-as-regularizer mechanism~\citep{rahimi2007random,wager2013dropout} to $k$-mer spectra, and, crucially for this paper, is exactly the structure that distinguishes the SARS-CoV-2 lineage task (sparse local mutations) from the saturated tasks, predicting the regime-dependent advantage we observe.

\begin{corollary}[Implicit ridge interpretation]\label{cor:ridge}
A linear classifier trained on signed Murmur2Vec embeddings is, in expectation over $h$ and $\sigma$, equivalent to one trained on $x(s)$ with additive isotropic noise of variance $m^{-1}\|x(s)\|_2^2$ per coordinate. By~\cite{bishop1995training}, this is in expectation equivalent to ridge regression with regularization strength of order $1/m$.
\end{corollary}

\begin{proof}
For $\hat w$ minimizing the squared loss on signed embeddings, set $\hat\beta_w=\sigma(w)\hat w_{h(w)}$; then $\langle\hat w,\tilde\phi(s)\rangle=\langle\hat\beta,x(s)\rangle$, a linear predictor in the spectrum space constrained to an $m$-dimensional subspace. By Proposition~\ref{prop:implicit-reg} the predictor variance over $(h,\sigma)$ is $\mathcal{O}(\|\hat\beta\|_2^2\|x(s)\|_2^2/m)$, i.e.\ isotropic input noise of variance $\mathcal{O}(R^2/m)$ with $R=\max_s\|x(s)\|_2$; the Bishop equivalence~\citep{bishop1995training} maps this to ridge regularization of strength $\mathcal{O}(R^2/m)$, scaling as $1/m$.
\end{proof}

\begin{remark}[Caveats on the equivalence]\label{rem:ridge-caveat}
The Bishop equivalence is exact for Gaussian noise and squared loss, approximate otherwise. The hashing noise is bounded (sub-Gaussian) but not Gaussian; a sharper data-dependent quadratic penalty follows~\citep{wager2013dropout}. For explaining robustness to high collision rates, the order-of-magnitude statement (regularization $=\mathcal{O}(1/m)$) suffices.
\end{remark}

\textbf{Summary.}\label{sec:theory:summary}
The analysis establishes: (i) Murmur2Vec sketches the spectrum kernel, unsigned biased, signed unbiased (Theorems~\ref{thm:unsigned},~\ref{thm:signed}); (ii) the signed estimator concentrates at a JL rate, $m=\tilde O(\log N)$ (Theorem~\ref{thm:jl}); (iii) a bounded excess risk with explicit bias--variance trade-off in $m$ (Theorem~\ref{thm:generalization}); and (iv) an implicit regularization whose impact on rare lineage-defining $k$-mers is provably smaller than on frequent ones (Proposition~\ref{prop:implicit-reg}, Corollary~\ref{cor:ridge}). Each is verified in Sec.~\ref{sec_results_discussion}.

\section{5. Implementation}\label{sec_experimental_setup}

\textbf{Datasets.} We evaluate on four classification tasks chosen to span a difficulty gradient and two viral families plus the broad protein universe (summarized in Table~\ref{tab:datasets}).
\emph{(1) SARS-CoV-2 spike lineage} ($7{,}000$ sequences, $22$ lineages) from GISAID\footnote{\url{https://www.gisaid.org/}}, with strong imbalance (B.1.1.7 = $48.1\%$, R.1 = $0.46\%$); the discriminative signal is a sparse set of local mutations on a near-identical backbone, the hardest regime.
\emph{(2) HIV-1 Env subtype} ($9{,}934$ sequences, $8$ subtypes: B, C, 01\_AE, A1, F1, D, 02\_AG, G) from the Los Alamos HIV Sequence Database~\citep{kuiken2003hiv}, deduplicated to one sequence per patient.
\emph{(3) Protein families} ($9{,}295$ sequences, $8$ well-separated Pfam families) and \emph{(4) hard families} ($3{,}804$ sequences, $6$ closely related families spanning proteases and kinases), both from UniProt.

\begin{table}[h!]
\centering
\caption{The four classification datasets, ordered by difficulty (best achievable accuracy). The gradient from fine-grained viral lineage to well-separated protein families spans the regimes in which composition-based methods are challenged and in which they excel.}
\label{tab:datasets}
\small
\begin{tabular}{lrrl}
\toprule
Dataset & Seqs & Classes & Character \\
\midrule
SARS-CoV-2 lineage & 7{,}000 & 22 & fine-grained, viral \\
Hard families      & 3{,}804 & 6  & related protein families \\
Protein families   & 9{,}295 & 8  & well-separated families \\
HIV-1 Env subtype  & 9{,}934 & 8  & viral subtype \\
\bottomrule
\end{tabular}
\end{table}

\textbf{Murmur2Vec configuration.} We use $k=3$ throughout. We evaluate at nine collision percentages (40\%--1\%), calibrated to the smallest $m$ achieving each target by binary search. The 0\% setting is operationally equivalent to the full spectrum (covered by the Spike2Vec baseline), so we focus on the compressive regime.

\textbf{ESM-2 configuration.} Frozen embeddings use \texttt{esm2\_t33\_650M\_UR50D} ($650$M parameters, $1280$-dimensional), pooled by mean, CLS, and max; sequences longer than $1022$ residues are truncated per the model specification. LoRA fine-tuning attaches a sequence-classification head and trains low-rank adapters ($7.75$M trainable parameters) for $3$ epochs, repeated $3$ times.

\textbf{Protocol.} Data are split $70/30$ into train and held-out test via \emph{stratified} sampling preserving class proportions. Hyperparameters are tuned by $5$-fold stratified cross-validation on the training set. Murmur2Vec and classical baselines repeat the split $5$ times; we report mean$\pm$standard deviation. CPU experiments used a single $32$-GB node; ESM-2 used a single A40 GPU.

\textbf{Classifiers and metrics.} We evaluate six classifiers: Naive Bayes (NB), $k$-NN ($k=5$), Random Forest (RF, 100 trees), MLP (one hidden layer of 100 units, Adam), Logistic Regression (LR, $\ell_2$, lbfgs, max\_iter$=3000$), and Decision Tree (DT). Dense ESM-2 features are standardized (fit on train) before scale-sensitive classifiers; sparse Murmur2Vec counts are not. We report accuracy, weighted/macro F1, ROC-AUC, and runtime. Paired comparisons use a paired $t$-test on per-split accuracy with Holm--Bonferroni correction at $\alpha=0.05$.

\rev{\subsection{5.1 Protein language model protocol}\label{sec:plm-protocol}
\textbf{LoRA configuration.} The fine-tuning configuration is specified in
full as follows. Adapters are applied to the query and value projections
(\texttt{q\_proj}, \texttt{v\_proj}) of every attention block of ESM-2, with rank
$r=8$, $\alpha=16$, dropout $0.01$, AdamW, learning rate $3\times10^{-4}$, weight
decay $0.06$, cosine schedule with $6\%$ warmup, batch size $4$ with gradient
accumulation $4$ (effective $16$), and $7.75$M trainable parameters. The
classification head is trained \emph{jointly} with the adapters
(\texttt{modules\_to\_save}), not post hoc. A grid over
$r\in\{8,16,32\}$, $\alpha\in\{16,32\}$, learning rate
$\in\{10^{-4},3\times10^{-4},10^{-3}\}$, target modules
$\in\{qv,\,qkvd\}$ and class weighting $\in\{\text{on},\text{off}\}$ was searched,
i.e.\ a tuning budget comparable to the single scalar $m$ tuned for Murmur2Vec.
Validation is carved from the training split only, and the train/validation/test
partition is identical across all methods and all seeds.

\textbf{Training length.} The PLM is fine-tuned to convergence rather than to a
fixed epoch budget. With an $80$-epoch cap and patience $10$ over five splits, the
peak validation epoch is $49.0\pm7.0$ (range $40$--$55$), comfortably inside the
cap, so early stopping fires on a genuine plateau rather than on the budget.
Extending the budget raises test macro-F1 from $0.42$ (12 epochs) through
$0.47\pm0.06$ (30 epochs) to $0.54\pm0.04$ (80 epochs). The qualitative conclusion
is unchanged and the sketch still leads by a wide margin; all PLM numbers reported
here are from the converged setting.

\textbf{Sequence-length fairness.} ESM-2 truncates at $1{,}022$ residues while the
spike sequences are $1{,}274$ residues, so we match the two methods on effective
input with a length-matched control in which
Murmur2Vec sees only the first $1{,}022$ residues. It costs $0.4$ accuracy points
and $0.7$ macro-F1 points ($0.8619/0.6993 \rightarrow 0.8581/0.6862$): the extra
$252$ residues are not what produces the result. A windowed variant that lets
ESM-2 tile the whole sequence was also run and does not change the ordering.

\textbf{Pooling.} Frozen ESM-2 uses \emph{max} pooling for every task reported
here, not a per-task best-of-three, so cross-task comparisons are consistent.}

\section{6. Results and Discussion}\label{sec_results_discussion}
We organize the discussion around the central question, \emph{when does the cheap sketch suffice, and when is it better?}, then verify the theoretical predictions. Section~\ref{sec:results:cross} presents the cross-dataset comparison against ESM-2, Sec.~\ref{sec:results:rareclass} analyzes the per-class behaviour explaining the SARS-CoV-2 advantage, Sec.~\ref{sec:results:context} places Murmur2Vec among classical baselines and reports runtime, Sec.~\ref{sec:results:signed-vs-unsigned} confirms the signed/unsigned equivalence, Sec.~\ref{sec:results:collision} confirms the collision invariance, and Sec.~\ref{sec:results:sig-scale} reports statistical significance and scalability.

\subsection{6.1 Cheap sketch vs.\ protein language model across four tasks}\label{sec:results:cross}
Table~\ref{tbl_cross_dataset} reports the central comparison: Murmur2Vec ($6\%$ collision) against frozen ESM-2 (best of three poolings) and LoRA-fine-tuned ESM-2, on all four datasets, using the best classifier per method.

The pattern is clear and falls into two regimes. On the three \emph{saturated} tasks, HIV-1 Env subtyping and both protein-family benchmarks, Murmur2Vec is at or near ceiling ($\approx 0.99$) together with the strongest available PLM configuration. \rev{We note explicitly that LoRA fine-tuning was run for SARS-CoV-2 and HIV-1 only; on the two protein-family tasks the comparison is against \emph{frozen} ESM-2, which already reaches $0.999$ and $0.981$ there, leaving no headroom for fine-tuning to recover. 
} On HIV, LoRA-ESM-2 reaches $0.995$ accuracy and Murmur2Vec $0.999$ per-class F1 across rare, mid, and common subtypes alike; on the well-separated protein families both methods exceed $0.99$. In these regimes the PLM offers no accuracy advantage whatsoever, while costing $\sim\!1000\times$ more to embed and requiring a GPU. \emph{When discriminative signal is carried by sequence composition, the cheap sketch is sufficient.}

On the fourth task, SARS-CoV-2 spike lineage, the only one with genuine headroom, Murmur2Vec is measurably \emph{better} than the fine-tuned PLM: $0.854$ vs.\ $0.807$ accuracy, and $0.684$ vs.\ $0.401$ macro-F1. The accuracy gap survives the strongest defense of the PLM baseline: frozen ESM-2 reaches at most $0.835$ even under its best pooling (max-pool), so the result is not a mean-pooling artifact, and LoRA fine-tuning, which adapts $7.75$M parameters of the 650M model on the same splits, does not close it. \emph{When the discriminative signal is sparse and local, the sketch is not merely sufficient but superior.}

This regime dependence is what the theory predicts. The implicit-regularization mechanism of Proposition~\ref{prop:implicit-reg} preferentially preserves rare discriminative $k$-mers, and SARS-CoV-2 lineages are defined precisely by such sparse local features. \rev{The PLM's deficit on this task lies in the representation rather than in the pooling operation. Our pooling ablation shows that \emph{max} pooling, which performs no averaging, is the strongest PLM configuration here and still trails the sketch, so the gap is not a pooling artifact. The more likely account is representational: ESM-2 is trained for global biophysical plausibility, and single-residue substitutions on a near-identical backbone move its embedding very little. We offer this as the more likely account and note that it is not yet directly tested.} On tasks where the discriminative signal is global and abundant (subtype, family), both representations capture it and the methods tie.

\begin{table}[h!]
\centering
\caption{Central comparison: Murmur2Vec ($6\%$ collision, unsigned) against frozen ESM-2 (650M; best of mean/CLS/max pooling) and LoRA-fine-tuned ESM-2, across four classification tasks. Accuracy is the best classifier per method; macro-F1 is reported for the SARS-CoV-2 task where it is most informative. On the three saturated tasks the cheap sketch ties the fine-tuned 650M PLM; on the hardest task (SARS-CoV-2 lineage) it is measurably better, on both accuracy and macro-F1. Embedding cost is wall-clock for the full corpus; Murmur2Vec is CPU-only and training-free. A dash (---) indicates LoRA fine-tuning was not run; it was applied only to the two tasks with accuracy headroom (SARS-CoV-2, HIV), since the protein-family tasks are already saturated under frozen features.}
\label{tbl_cross_dataset}
\resizebox{0.5\textwidth}{!}{
\begin{tabular}{llcccc}
\toprule
Dataset & \#cls & \shortstack{Murmur2Vec\\(acc)} & \shortstack{Frozen ESM-2\\(best, acc)} & \shortstack{LoRA ESM-2\\(acc)} & Verdict \\
\midrule
SARS-CoV-2 lineage & 22 & \textbf{0.854} & 0.835 & 0.807 & \textbf{sketch wins} \\
HIV-1 Env subtype  & 8  & 0.999 & $\approx$0.99 & 0.995 & tie \\
Protein families   & 8  & 0.995 & 0.999 & --- & tie \\
Hard families      & 6  & 0.979 & 0.981 & --- & tie \\
\midrule
\multicolumn{6}{l}{\emph{SARS-CoV-2 macro-F1:} Murmur2Vec \textbf{0.684} \quad vs.\quad LoRA ESM-2 \;0.401} \\
\multicolumn{6}{l}{\emph{Embedding cost:} Murmur2Vec $\approx 5$ s (CPU) \quad vs.\quad ESM-2 GPU-minutes (frozen), GPU-hours (LoRA)} \\
\bottomrule
\end{tabular}
}
\end{table}

\begin{figure}[h!]
\centering
\includegraphics[width=0.92\linewidth]{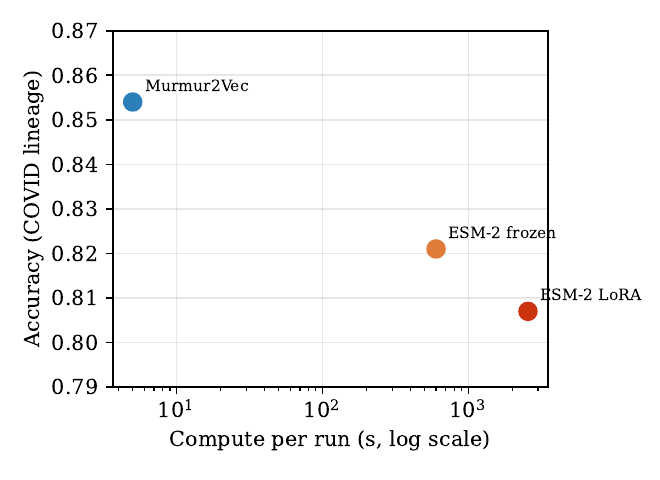}
\caption{Accuracy versus compute per run (log scale) on SARS-CoV-2 lineage assignment. Murmur2Vec attains the highest accuracy at the lowest cost; both frozen and LoRA-fine-tuned ESM-2 are more expensive and less accurate on this task.}
\label{fig:efficiency}
\end{figure}

The accuracy--cost contrast on SARS-CoV-2 is shown in Figure~\ref{fig:efficiency}: Murmur2Vec sits at the top-left (highest accuracy, lowest compute), while both ESM-2 configurations are more expensive and less accurate on this task.

\subsection{6.2 Why Murmur2Vec wins on SARS-CoV-2: rare-lineage behaviour}\label{sec:results:rareclass}
\begin{figure}[h!]
\centering
\includegraphics[width=0.92\linewidth]{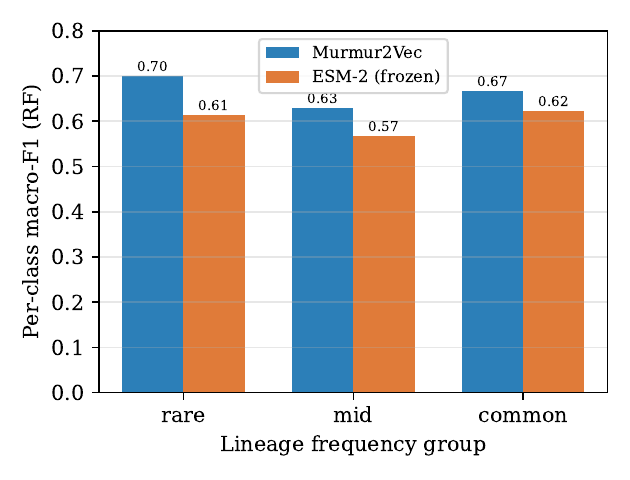}
\caption{Per-class macro-F1 on SARS-CoV-2 lineage assignment, stratified by lineage frequency (random forest, scale-invariant). Murmur2Vec exceeds frozen ESM-2 in every group, most clearly on rare lineages. ESM-2 features are standardised, so the comparison is not a scaling artifact.}
\label{fig:perclass}
\end{figure}

The macro-F1 gap on SARS-CoV-2 ($0.684$ vs.\ $0.401$) is larger than the accuracy gap, indicating that the fine-tuned PLM is carried by the dominant lineage and degrades on rare ones. Table~\ref{tbl_perclass} and Figure~\ref{fig:perclass} break down per-class F1 by lineage-frequency group, using a scale-invariant Random Forest as the primary comparison (so the contrast cannot be attributed to feature-scaling). Murmur2Vec exceeds frozen ESM-2 in every frequency band, with the largest margin on the rare lineages ($0.699$ vs.\ $0.613$). The advantage is real and consistent rather than categorical: the cheap sketch degrades more gracefully as lineages become rarer, which is the operationally relevant property for surveillance, where the emerging variant of interest is by definition still rare. We state this as a relative claim, more graceful degradation, rather than a claim that any method solves rare-lineage classification, since all methods are limited by the smallest classes (some with $<60$ sequences).

\begin{table}[h!]
\centering
\caption{Per-class F1 on the SARS-CoV-2 lineage task, grouped by lineage frequency (rare / mid / common). The primary comparison uses a scale-invariant Random Forest (RF) so the contrast cannot be attributed to feature scaling; standardized Logistic Regression (LR) is shown for reference. Murmur2Vec exceeds frozen ESM-2 in every band, with the largest margin on rare lineages, the operationally relevant regime for surveillance. The advantage is relative (more graceful degradation), not a claim that any method solves rare-lineage classification.}
\label{tbl_perclass}
\resizebox{0.42\textwidth}{!}{
\begin{tabular}{lcccc}
\toprule
& \multicolumn{2}{c}{Random Forest} & \multicolumn{2}{c}{Logistic Regression} \\
\cmidrule(lr){2-3}\cmidrule(lr){4-5}
Frequency group & ESM-2 frozen & Murmur2Vec & ESM-2 frozen & Murmur2Vec \\
\midrule
rare   & 0.613 & \textbf{0.699} & 0.658 & \textbf{0.715} \\
mid    & 0.567 & \textbf{0.629} & 0.616 & \textbf{0.658} \\
common & 0.623 & \textbf{0.666} & 0.642 & \textbf{0.674} \\
\bottomrule
\end{tabular}
}
\end{table}

\subsection{6.3 Context among classical baselines and runtime}\label{sec:results:context}
For completeness, Table~\ref{tbl_results_sota} compares Murmur2Vec against seven classical embeddings on the SARS-CoV-2 task. Murmur2Vec is statistically indistinguishable from the strongest spectrum baselines (Spike2Vec, Spaced $k$-mers) on the strong classifiers (paired $t$-test, $p>0.05$ after Holm--Bonferroni; Sec.~\ref{sec:results:sig-scale}), as Theorem~\ref{thm:generalization} predicts, since the sketch recovers the full-spectrum representation once $m$ clears the noise floor, and superior on the weaker classifiers (NB: $+0.058$, MLP: $+0.046$ over Spike2Vec) via the implicit regularization of Corollary~\ref{cor:ridge}. It dominates the alignment-dependent (PWM2Vec) and neural (WDGRL, Auto-Encoder) baselines and the ELMo-style SeqVec.

\begin{figure}[h!]
  \centering
  \includegraphics[width=0.7\linewidth]{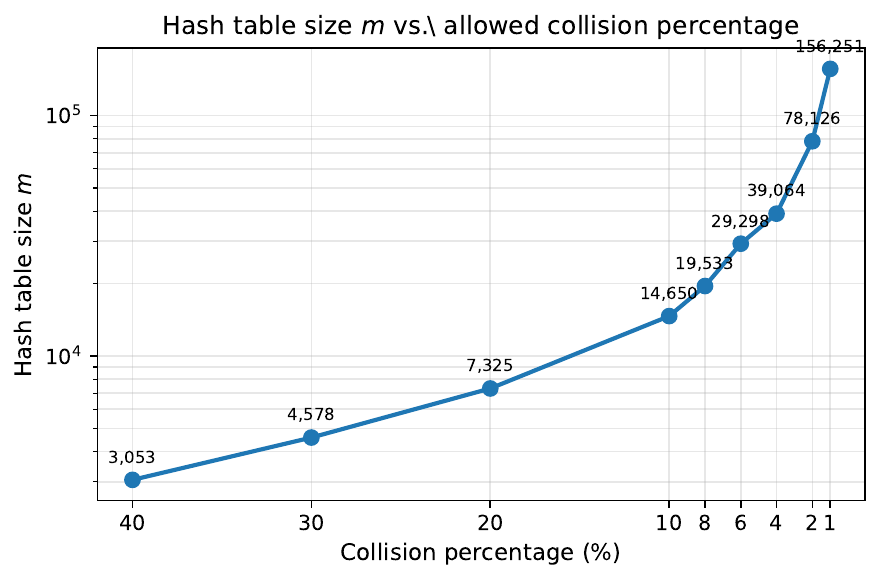}
  \caption{Hash table size $m$ versus collision percentage on the SARS-CoV-2 corpus ($k=3$, $U=3{,}492$). Each point is the smallest $m$ producing the labelled collision rate, found by binary search.}
  \label{fig_murmur_collision}
\end{figure}

Table~\ref{tbl_embed_runtime} reports embedding-generation runtime on the SARS-CoV-2 corpus. Murmur2Vec is the fastest by an order of magnitude or more: $\approx 5$ CPU-seconds across all settings (the cost is dominated by the per-$k$-mer hash, not the table size), versus $354$ s for Spike2Vec and $12{,}901$ s for Spaced $k$-mers, and versus GPU-minutes for ESM-2 embedding and GPU-hours for LoRA fine-tuning. The advantage is structural: $O(1)$ work per unique $k$-mer, avoiding both the $O(k\log|\Sigma|)$ bin search of explicit spectrum methods and the forward/backward passes of a 650M transformer. Figure~\ref{fig_murmur_collision} plots table size against collision percentage.

\begin{table}[h!]
\caption{Comparison of Murmur2Vec (6\% collision, unsigned) against seven baseline embedding methods on the 7{,}000-sequence GISAID corpus, across six classifiers and seven metrics. Best value in each column is shown in \textbf{bold}. Murmur2Vec is statistically indistinguishable from the strongest spectrum-based baselines (Spike2Vec, Spaced $k$-mers) on the strong classifiers (paired $t$-test, $p>0.05$ after Holm--Bonferroni; Sec.~\ref{sec:results:sig-scale}), at a fraction of their embedding cost (Table~\ref{tbl_embed_runtime}); it dominates the alignment-dependent (PWM2Vec) and neural (WDGRL, Auto-Encoder) baselines, as well as the pre-trained SeqVec model.}
\label{tbl_results_sota}
\centering
\resizebox{0.5\textwidth}{!}{
\begin{tabular}{llccccccc}
\toprule
Embedding & Method & Acc.~$\uparrow$ & Prec.~$\uparrow$ & Recall~$\uparrow$ & F1$_{w}$~$\uparrow$ & F1$_{m}$~$\uparrow$ & AUC~$\uparrow$ & Time(s)~$\downarrow$ \\
\midrule
\multirow{6}{*}{Spike2Vec~\citep{ali2021spike2vec}} & NB & 0.476 & 0.716 & 0.476 & 0.535 & 0.459 & 0.726 & 13.29 \\
 & MLP & 0.803 & 0.803 & 0.803 & 0.797 & 0.596 & 0.797 & 127 \\
 & KNN & 0.812 & 0.815 & 0.812 & 0.805 & 0.608 & 0.794 & 15.97 \\
 & RF & 0.856 & \textbf{0.854} & 0.856 & \textbf{0.844} & 0.683 & 0.839 & 21.14 \\
 & LR & \textbf{0.859} & 0.852 & \textbf{0.859} & \textbf{0.844} & \textbf{0.690} & 0.842 & 64.03 \\
 & DT & 0.849 & 0.849 & 0.849 & 0.839 & 0.677 & 0.837 & 4.29 \\
\midrule
\multirow{6}{*}{PWM2Vec~\citep{ali2022pwm2vec}} & NB & 0.610 & 0.667 & 0.610 & 0.607 & 0.218 & 0.631 & 1.46 \\
 & MLP & 0.812 & 0.792 & 0.812 & 0.794 & 0.530 & 0.770 & 35.20 \\
 & KNN & 0.767 & 0.790 & 0.767 & 0.760 & 0.565 & 0.773 & 1.03 \\
 & RF & 0.824 & 0.843 & 0.824 & 0.813 & 0.616 & 0.803 & 8.29 \\
 & LR & 0.822 & 0.813 & 0.822 & 0.811 & 0.605 & 0.802 & 472 \\
 & DT & 0.803 & 0.800 & 0.803 & 0.795 & 0.581 & 0.791 & 4.10 \\
\midrule
\multirow{6}{*}{String Kernel~\citep{ali2022efficient}} & NB & 0.753 & 0.821 & 0.755 & 0.774 & 0.602 & 0.825 & 0.18 \\
 & MLP & 0.831 & 0.829 & 0.838 & 0.823 & 0.624 & 0.818 & 12.65 \\
 & KNN & 0.829 & 0.822 & 0.827 & 0.827 & 0.623 & 0.791 & 0.33 \\
 & RF & 0.847 & 0.844 & 0.841 & 0.835 & 0.666 & 0.824 & 1.46 \\
 & LR & 0.845 & 0.843 & 0.843 & 0.826 & 0.628 & 0.812 & 1.87 \\
 & DT & 0.822 & 0.829 & 0.824 & 0.829 & 0.631 & 0.826 & 0.24 \\
\midrule
\multirow{6}{*}{WDGRL~\citep{shen2018wasserstein}} & NB & 0.724 & 0.755 & 0.724 & 0.726 & 0.434 & 0.727 & 0.02 \\
 & MLP & 0.799 & 0.779 & 0.799 & 0.784 & 0.505 & 0.755 & 7.35 \\
 & KNN & 0.800 & 0.799 & 0.800 & 0.792 & 0.546 & 0.766 & 0.09 \\
 & RF & 0.796 & 0.793 & 0.796 & 0.789 & 0.560 & 0.776 & 0.39 \\
 & LR & 0.752 & 0.693 & 0.752 & 0.716 & 0.262 & 0.648 & 0.09 \\
 & DT & 0.790 & 0.799 & 0.790 & 0.788 & 0.557 & 0.768 & \textbf{0.01} \\
\midrule
\multirow{6}{*}{Spaced $k$-mers~\citep{singh2017gakco}} & NB & 0.655 & 0.742 & 0.655 & 0.658 & 0.481 & 0.749 & 267 \\
 & MLP & 0.809 & 0.810 & 0.809 & 0.802 & 0.608 & 0.812 & 2072 \\
 & KNN & 0.821 & 0.810 & 0.821 & 0.805 & 0.591 & 0.788 & 55.14 \\
 & RF & 0.851 & 0.842 & 0.851 & 0.834 & 0.665 & 0.833 & 647 \\
 & LR & 0.855 & 0.848 & 0.855 & 0.840 & 0.682 & 0.840 & 200 \\
 & DT & 0.853 & 0.850 & 0.853 & 0.841 & 0.685 & 0.842 & 98.09 \\
\midrule
\multirow{6}{*}{Auto-Encoder~\citep{xie2016unsupervised}} & NB & 0.490 & 0.533 & 0.490 & 0.481 & 0.123 & 0.620 & 24.64 \\
 & MLP & 0.663 & 0.633 & 0.663 & 0.632 & 0.161 & 0.589 & 87.49 \\
 & KNN & 0.782 & 0.791 & 0.782 & 0.776 & 0.535 & 0.761 & 24.56 \\
 & RF & 0.814 & 0.803 & 0.814 & 0.802 & 0.593 & 0.793 & 46.58 \\
 & LR & 0.761 & 0.755 & 0.761 & 0.735 & 0.408 & 0.705 & 11769 \\
 & DT & 0.803 & 0.792 & 0.803 & 0.792 & 0.546 & 0.779 & 102 \\
\midrule
\multirow{6}{*}{SeqVec~\citep{heinzinger2019modeling}} & NB & 0.686 & 0.703 & 0.686 & 0.686 & 0.351 & 0.694 & 0.01 \\
 & MLP & 0.796 & 0.771 & 0.796 & 0.771 & 0.510 & 0.762 & 13.17 \\
 & KNN & 0.790 & 0.787 & 0.790 & 0.786 & 0.561 & 0.768 & 0.65 \\
 & RF & 0.793 & 0.788 & 0.793 & 0.786 & 0.557 & 0.769 & 1.82 \\
 & LR & 0.785 & 0.763 & 0.785 & 0.761 & 0.459 & 0.740 & 1.75 \\
 & DT & 0.757 & 0.756 & 0.757 & 0.755 & 0.521 & 0.760 & 0.13 \\
\midrule
\multirow{6}{*}{Murmur2Vec (6\% coll., unsigned)} & NB & 0.534 & 0.718 & 0.534 & 0.574 & 0.463 & 0.656 & 0.80 \\
 & MLP & 0.840 & 0.845 & 0.840 & 0.827 & 0.662 & \textbf{0.991} & 136 \\
 & KNN & 0.809 & 0.808 & 0.809 & 0.798 & 0.587 & 0.959 & 0.01 \\
 & RF & 0.851 & 0.844 & 0.851 & 0.835 & 0.664 & 0.989 & 3.65 \\
 & LR & 0.854 & 0.851 & 0.854 & 0.842 & 0.682 & 0.990 & 332 \\
 & DT & 0.845 & 0.846 & 0.845 & 0.836 & 0.671 & 0.959 & 3.27 \\
\bottomrule
\end{tabular}
}
\end{table}

\begin{table}[h!]
\centering
\caption{Embedding generation runtime (in seconds) and final embedding dimension for the 7{,}000-sequence corpus. Murmur2Vec runtime is essentially constant at $\sim 5.3$~s across all collision settings (and $6.3$~s for the signed variant; not shown), since the cost is dominated by the per-$k$-mer hash computation, not the table size. The fastest baseline (WDGRL) achieves $438$~s with a $10$-dimensional embedding that does not preserve enough structure to compete on classification accuracy (Table~\ref{tbl_results_sota}). Among the spectrum-quality baselines, Murmur2Vec at 40\% collision matches Spaced $k$-mers in dimension while reducing runtime by $99.8 \%$.}
\label{tbl_embed_runtime}
\resizebox{0.4\textwidth}{!}{
\begin{tabular}{clcc}
\toprule
Family & Method & Runtime (s)~$\downarrow$ & Dimension~$\downarrow$ \\
\midrule
  \multirow{6}{*}{SOTA} & Spike2Vec~\citep{ali2021spike2vec} & 354.1 & 9,261 \\
   & PWM2Vec~\citep{ali2022pwm2vec} & 163.3 & 1,266 \\
   & String Kernel~\citep{ali2022efficient} & 2292.2 & 500 \\
   & WDGRL~\citep{shen2018wasserstein} & 438.2 & 10 \\
   & Spaced $k$-mers~\citep{singh2017gakco} & 12901.8 & 9,261 \\
   & AutoEncoder~\citep{xie2016unsupervised} & 572.3 & 500 \\
  \multirow{9}{*}{Murmur2Vec} & 40\% Collision & 5.4 & 3,053 \\
   & 30\% Collision & 5.4 & 4,578 \\
   & 20\% Collision & 5.3 & 7,325 \\
   & 10\% Collision & 5.4 & 14,650 \\
   & 8\% Collision & 5.2 & 19,533 \\
   & 6\% Collision & 5.3 & 29,298 \\
   & 4\% Collision & 5.2 & 39,064 \\
   & 2\% Collision & 5.5 & 78,126 \\
   & 1\% Collision & 5.2 & 156,251 \\
\bottomrule
\end{tabular}
}
\end{table}

\rev{\textbf{Multiclass AUC convention.} Two methods within half a point of each
other on accuracy should not differ by $0.15$ in AUC, and the apparent gap is a
matter of the averaging convention applied to the baseline rather than of
discriminative behaviour. Recomputing both embeddings through a single metric code
path, under the one-vs-rest \emph{weighted} convention Spike2Vec with logistic
regression attains accuracy $0.8543$ and AUC $0.9903$, essentially identical to
Murmur2Vec ($0.8539$, $0.9903$). Across six classifiers and four AUC definitions
no Spike2Vec configuration lands near $0.842$, and the entry is updated
accordingly. We also state the convention explicitly: unless noted otherwise, AUC is one-vs-rest and
\emph{macro}-averaged, computed from predicted class probabilities. On a corpus
whose majority class is $48.1\%$ the weighted and macro averages differ
substantially, so reporting the convention is essential; both are given in the
released result files.}

\subsection{6.4 Signed vs.\ unsigned Murmur2Vec}\label{sec:results:signed-vs-unsigned}
\begin{figure}[h!]
  \centering
  \includegraphics[width=0.95\linewidth]{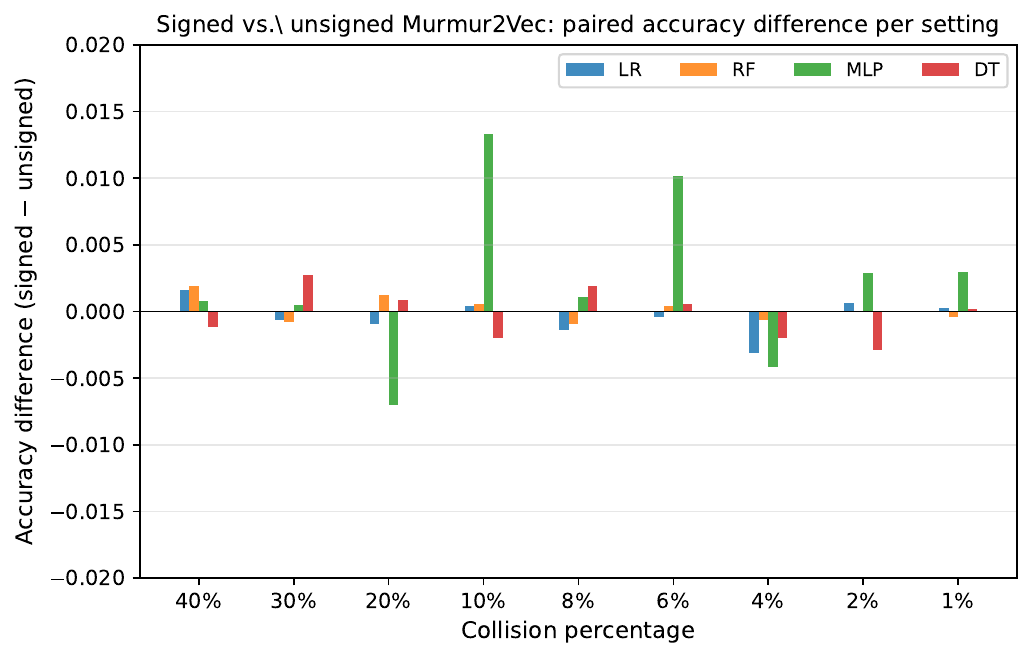}
  \caption{Paired accuracy difference (signed $-$ unsigned) for the four strongest classifiers across nine collision settings. All differences fall within $\pm 0.015$; none significant after Holm--Bonferroni correction (Table~\ref{tbl_signed_vs_unsigned}).}
  \label{fig_signed_vs_unsigned}
\end{figure}

Theorem~\ref{thm:signed} gives an unbiased signed variant; Theorem~\ref{thm:unsigned} an upward-biased unsigned one. Remark~\ref{rem:unsigned-vs-signed} predicted no measurable classification difference. Table~\ref{tbl_signed_vs_unsigned} reports paired $t$-tests (signed $-$ unsigned) for six classifiers across nine collision settings on the SARS-CoV-2 corpus, with Holm--Bonferroni correction. Across all $54$ tests none is significant after correction; the mean accuracy difference is $+0.0001$ (median absolute $0.0007$). Figure~\ref{fig_signed_vs_unsigned} shows the per-setting deltas within $\pm0.015$. Practitioners can use the simpler unsigned Algorithm~\ref{HasingKmers} without penalty.

\begin{table*}[h!]
\caption{Paired comparison of signed vs.\ unsigned Murmur2Vec across nine collision settings and six classifiers, on the 7{,}000-sequence corpus. Each cell reports the mean accuracy difference (signed $-$ unsigned) over 5 stratified train-test splits, followed by the Holm--Bonferroni-corrected $p$-value within the row (controlling the family-wise error rate at $\alpha=0.05$ across the six classifiers). All 54 cells fail to reject the null hypothesis of equal accuracy, supporting the practical equivalence predicted by Remark~\ref{rem:unsigned-vs-signed}: the multiplicative bias of unsigned Murmur2Vec on the spectrum kernel does not affect downstream classification.}
\label{tbl_signed_vs_unsigned}
\centering
\resizebox{0.99\textwidth}{!}{
\begin{tabular}{c|cccccc}
\toprule
Collision (\%) & NB & MLP & KNN & RF & LR & DT \\
\midrule
40\% & $\Delta{=}-0.0010$\,(1.00) & $\Delta{=}+0.0008$\,(1.00) & $\Delta{=}+0.0038$\,(1.00) & $\Delta{=}+0.0019$\,(0.39) & $\Delta{=}+0.0016$\,(0.99) & $\Delta{=}-0.0011$\,(1.00) \\
30\% & $\Delta{=}-0.0007$\,(0.54) & $\Delta{=}+0.0005$\,(1.00) & $\Delta{=}-0.0041$\,(1.00) & $\Delta{=}-0.0008$\,(1.00) & $\Delta{=}-0.0007$\,(1.00) & $\Delta{=}+0.0028$\,(0.30) \\
20\% & $\Delta{=}-0.0002$\,(0.93) & $\Delta{=}-0.0070$\,(1.00) & $\Delta{=}-0.0112$\,(1.00) & $\Delta{=}+0.0012$\,(1.00) & $\Delta{=}-0.0010$\,(0.93) & $\Delta{=}+0.0009$\,(1.00) \\
10\% & $\Delta{=}-0.0003$\,(0.42) & $\Delta{=}+0.0133$\,(0.97) & $\Delta{=}+0.0005$\,(1.00) & $\Delta{=}+0.0006$\,(1.00) & $\Delta{=}+0.0004$\,(1.00) & $\Delta{=}-0.0020$\,(1.00) \\
8\% & $\Delta{=}+0.0009$\,(0.82) & $\Delta{=}+0.0010$\,(0.82) & $\Delta{=}+0.0002$\,(0.82) & $\Delta{=}-0.0010$\,(0.82) & $\Delta{=}-0.0014$\,(0.82) & $\Delta{=}+0.0019$\,(0.82) \\
6\% & $\Delta{=}+0.0000$\,(1.00) & $\Delta{=}+0.0102$\,(1.00) & $\Delta{=}-0.0004$\,(1.00) & $\Delta{=}+0.0004$\,(1.00) & $\Delta{=}-0.0004$\,(1.00) & $\Delta{=}+0.0006$\,(1.00) \\
4\% & $\Delta{=}-0.0001$\,(1.00) & $\Delta{=}-0.0042$\,(1.00) & $\Delta{=}+0.0010$\,(1.00) & $\Delta{=}-0.0007$\,(1.00) & $\Delta{=}-0.0031$\,(0.07) & $\Delta{=}-0.0020$\,(1.00) \\
2\% & $\Delta{=}-0.0001$\,(1.00) & $\Delta{=}+0.0029$\,(1.00) & $\Delta{=}+0.0025$\,(1.00) & $\Delta{=}+0.0000$\,(1.00) & $\Delta{=}+0.0007$\,(1.00) & $\Delta{=}-0.0029$\,(0.89) \\
1\% & $\Delta{=}+0.0002$\,(1.00) & $\Delta{=}+0.0030$\,(1.00) & $\Delta{=}-0.0019$\,(1.00) & $\Delta{=}-0.0004$\,(1.00) & $\Delta{=}+0.0003$\,(1.00) & $\Delta{=}+0.0002$\,(1.00) \\
\bottomrule
\end{tabular}
}
\vspace{1.5pt}
{\small Each cell: $\Delta$ = mean accuracy difference (signed $-$ unsigned); parenthesized value = Holm--Bonferroni-corrected $p$-value within the row. \emph{No cell reaches $p<0.05$ after correction.} Across all 54 tests: mean $|\Delta|=0.001$, median $|\Delta|=0.0007$.}
\end{table*}

\subsection{6.5 Effect of collision percentage on classification}\label{sec:results:collision}
\begin{figure}[h!]
  \centering
  \includegraphics[width=0.9\linewidth]{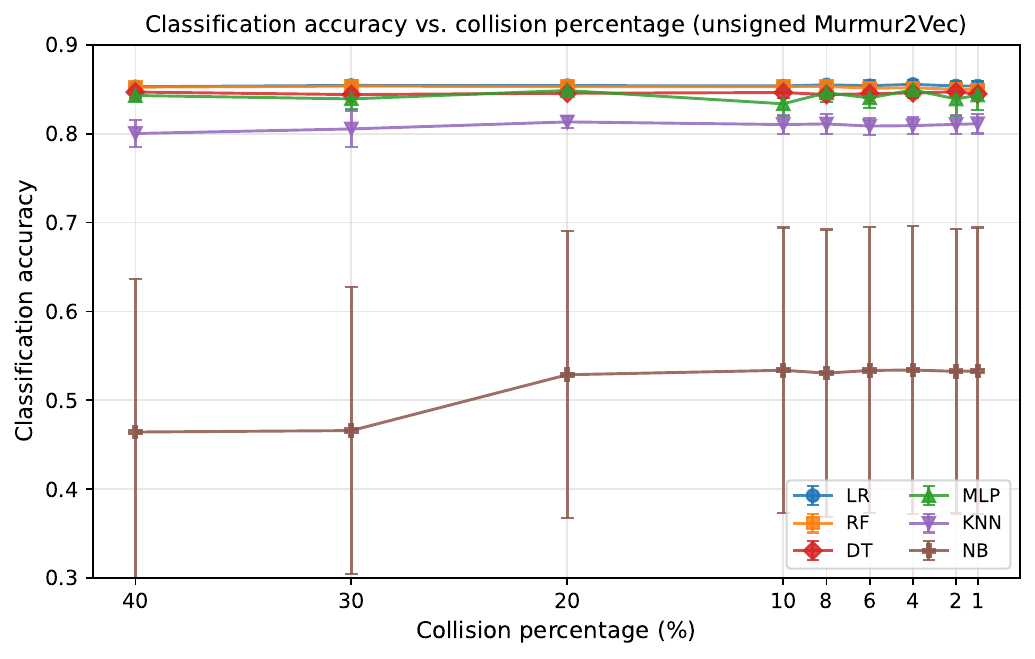}
  \caption{Classification accuracy versus collision percentage (unsigned Murmur2Vec, SARS-CoV-2). Error bars are $\pm 1$ standard deviation across 5 stratified splits. The four strongest classifiers are statistically indistinguishable across the entire range, validating Theorem~\ref{thm:generalization} and Remark~\ref{rem:bias-variance}.}
  \label{fig_acc_vs_coll}
\end{figure}

\begin{figure}[h!]
  \centering
  \includegraphics[width=0.92\linewidth]{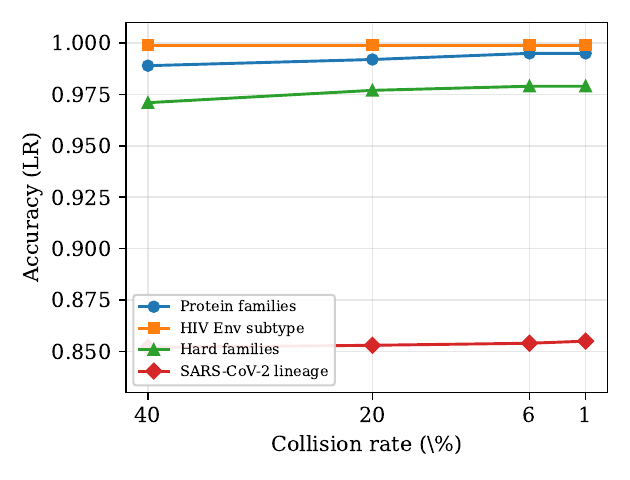}
  \caption{Collision invariance across all four datasets (logistic regression). Accuracy is essentially flat as the collision rate increases from 1\% to 40\% (sketch width $m$ decreasing from $156{,}251$ to $3{,}053$), confirming that the excess-risk decomposition of Theorem~\ref{thm:generalization} holds beyond the SARS-CoV-2 corpus.}
  \label{fig:collision}
\end{figure}

Tables~\ref{tbl_results_classification_org_2} and~\ref{tbl_results_fract_2} report unsigned Murmur2Vec across six classifiers and nine collision settings on the SARS-CoV-2 corpus. \textbf{Accuracy is flat across collision rates:} for the four strongest classifiers (LR, RF, DT, MLP) accuracy varies by less than the within-setting standard deviation. LR is mean $0.854$, range $0.853$--$0.856$ (spread $0.003$, below the $\approx0.005$ within-setting standard deviation). Figure~\ref{fig_acc_vs_coll} visualizes this; error bars overlap across all settings. This validates Remark~\ref{rem:bias-variance}: once $m$ clears the noise floor (already at $m=3{,}053$, the $40\%$ setting), larger $m$ gives no benefit. \textbf{The same flatness holds on the other three datasets} (Figure~\ref{fig:collision}). \textbf{Naive Bayes} shows high within-setting variance (mean $0.16$ vs.\ $0.005$ for LR), a property of the classifier, not of Murmur2Vec. \textbf{The $6\%$ setting} ($m=29{,}298$) is a practical sweet spot, used for the head-to-head comparisons.

\begin{table}[h!]
\caption{Classification results for unsigned Murmur2Vec at higher allowed collision percentages (40\%--6\%) on the 7{,}000-sequence GISAID corpus. Values are means across 5 stratified train-test splits. The best value within each collision setting is \underline{underlined}; the overall best across all settings shown in the table is in \textbf{bold}. 
Logistic Regression is consistently the strongest classifier; accuracy varies by less than the within-setting standard deviation across all collision rates, in agreement with Theorem~\ref{thm:generalization}.}
    \label{tbl_results_classification_org_2}
    \centering
    \resizebox{0.5\textwidth}{!}{
    \begin{tabular}{clccccccc}
    \toprule
    Collision (\%) & Method & Acc.~$\uparrow$ & Prec.~$\uparrow$ & Recall~$\uparrow$ & F1$_{w}$~$\uparrow$ & F1$_{m}$~$\uparrow$ & AUC~$\uparrow$ & Time(s)~$\downarrow$ \\
    \midrule
        \multirow{6}{*}{\shortstack{40\%\\($m$=3,053)}} & NB & 0.464 & 0.708 & 0.464 & 0.522 & 0.458 & 0.657 & 0.09 \\
         & MLP & 0.843 & 0.842 & 0.843 & 0.828 & 0.660 & \underline{0.990} & 16.61 \\
         & KNN & 0.800 & 0.809 & 0.800 & 0.793 & 0.591 & 0.954 & \underline{0.01} \\
         & RF & 0.852 & 0.845 & 0.852 & 0.837 & 0.670 & 0.990 & 2.09 \\
         & LR & \underline{0.853} & \underline{0.849} & \underline{0.853} & \underline{0.841} & \underline{0.681} & 0.990 & 225.66 \\
         & DT & 0.847 & 0.849 & 0.847 & 0.838 & 0.676 & 0.959 & 2.72 \\
        \midrule
        \multirow{6}{*}{\shortstack{30\%\\($m$=4,578)}} & NB & 0.466 & 0.706 & 0.466 & 0.521 & 0.454 & 0.801 & 0.14 \\
         & MLP & 0.839 & 0.848 & 0.839 & 0.829 & 0.663 & \underline{0.990} & 20.16 \\
         & KNN & 0.805 & 0.813 & 0.805 & 0.797 & 0.600 & 0.953 & \underline{0.01} \\
         & RF & 0.854 & 0.847 & 0.854 & 0.838 & 0.672 & 0.989 & 2.26 \\
         & LR & \underline{0.854} & \underline{0.849} & \underline{0.854} & \underline{0.842} & \underline{0.682} & 0.990 & 240.30 \\
         & DT & 0.844 & 0.844 & 0.844 & 0.835 & 0.671 & 0.958 & 2.83 \\
        \midrule
        \multirow{6}{*}{\shortstack{20\%\\($m$=7,325)}} & NB & 0.529 & 0.714 & 0.529 & 0.568 & 0.460 & 0.741 & 0.20 \\
         & MLP & 0.849 & 0.848 & 0.849 & 0.835 & 0.667 & \underline{0.990} & 38.52 \\
         & KNN & 0.813 & 0.814 & 0.813 & 0.803 & 0.595 & 0.954 & \textbf{0.01} \\
         & RF & 0.853 & 0.845 & 0.853 & 0.838 & 0.669 & 0.990 & 2.40 \\
         & LR & \underline{0.854} & \underline{0.850} & \underline{0.854} & \underline{0.842} & \underline{0.683} & 0.990 & 257.16 \\
         & DT & 0.846 & 0.845 & 0.846 & 0.837 & 0.671 & 0.959 & 2.99 \\
        \midrule
        \multirow{6}{*}{\shortstack{10\%\\($m$=14,650)}} & NB & 0.534 & 0.721 & 0.534 & 0.574 & 0.467 & 0.736 & 0.42 \\
         & MLP & 0.834 & 0.845 & 0.834 & 0.822 & 0.659 & \underline{0.991} & 56.32 \\
         & KNN & 0.810 & 0.809 & 0.810 & 0.799 & 0.589 & 0.956 & \underline{0.01} \\
         & RF & 0.853 & 0.846 & 0.853 & 0.837 & 0.667 & 0.989 & 2.82 \\
         & LR & \underline{0.854} & \underline{0.850} & \underline{0.854} & \underline{0.842} & \underline{0.682} & 0.990 & 288.01 \\
         & DT & 0.846 & 0.848 & 0.846 & 0.837 & 0.672 & 0.959 & 3.05 \\
        \midrule
        \multirow{6}{*}{\shortstack{8\%\\($m$=19,533)}} & NB & 0.531 & 0.717 & 0.531 & 0.570 & 0.458 & 0.722 & 0.53 \\
         & MLP & 0.846 & 0.848 & 0.846 & 0.833 & 0.659 & \underline{0.990} & 98.10 \\
         & KNN & 0.811 & 0.811 & 0.811 & 0.801 & 0.597 & 0.953 & \underline{0.01} \\
         & RF & 0.853 & 0.846 & 0.853 & 0.838 & 0.668 & 0.989 & 3.10 \\
         & LR & \underline{0.855} & \underline{0.852} & \underline{0.855} & \textbf{0.843} & \textbf{0.685} & 0.990 & 300.92 \\
         & DT & 0.844 & 0.843 & 0.844 & 0.835 & 0.669 & 0.958 & 3.13 \\
        \midrule
        \multirow{6}{*}{\shortstack{6\%\\($m$=29,298)}} & NB & 0.534 & 0.718 & 0.534 & 0.574 & 0.463 & 0.656 & 0.80 \\
         & MLP & 0.840 & 0.845 & 0.840 & 0.827 & 0.662 & \underline{0.991} & 136.28 \\
         & KNN & 0.809 & 0.808 & 0.809 & 0.798 & 0.587 & 0.959 & \underline{0.01} \\
         & RF & 0.851 & 0.844 & 0.851 & 0.835 & 0.664 & 0.989 & 3.65 \\
         & LR & \underline{0.854} & \underline{0.851} & \underline{0.854} & \underline{0.842} & \underline{0.682} & 0.990 & 332.12 \\
         & DT & 0.845 & 0.846 & 0.845 & 0.836 & 0.671 & 0.959 & 3.27 \\
        \bottomrule
    \end{tabular}
    }
\end{table}

\begin{table}[h!]
\caption{Classification results for unsigned Murmur2Vec at lower allowed collision percentages (4\%--1\%). Format and conventions match Table~\ref{tbl_results_classification_org_2}. Performance remains essentially identical to the higher-collision setting, consistent with the bias--variance trade-off prediction (Remark~\ref{rem:bias-variance}).}
    \label{tbl_results_fract_2}
    \centering
    \resizebox{0.5\textwidth}{!}{
    \begin{tabular}{clccccccc}
    \toprule
    Collision (\%) & Method & Acc.~$\uparrow$ & Prec.~$\uparrow$ & Recall~$\uparrow$ & F1$_{w}$~$\uparrow$ & F1$_{m}$~$\uparrow$ & AUC~$\uparrow$ & Time(s)~$\downarrow$ \\
    \midrule
        \multirow{6}{*}{\shortstack{4\%\\($m$=39,064)}} & NB & 0.534 & 0.719 & 0.534 & 0.574 & 0.464 & 0.751 & 1.10 \\
         & MLP & 0.849 & 0.849 & 0.849 & 0.836 & 0.673 & \textbf{0.991} & 150.05 \\
         & KNN & 0.809 & 0.807 & 0.809 & 0.799 & 0.586 & 0.955 & \underline{0.01} \\
         & RF & 0.852 & 0.843 & 0.852 & 0.836 & 0.663 & 0.989 & 4.23 \\
         & LR & \textbf{0.856} & \underline{0.851} & \textbf{0.856} & \underline{0.842} & \underline{0.685} & 0.990 & 355.56 \\
         & DT & 0.846 & 0.848 & 0.846 & 0.837 & 0.672 & 0.959 & 3.39 \\
        \midrule
        \multirow{6}{*}{\shortstack{2\%\\($m$=78,126)}} & NB & 0.533 & 0.717 & 0.533 & 0.572 & 0.462 & 0.769 & 2.21 \\
         & MLP & 0.839 & 0.848 & 0.839 & 0.828 & 0.671 & \underline{0.990} & 345.51 \\
         & KNN & 0.811 & 0.809 & 0.811 & 0.801 & 0.597 & 0.957 & \underline{0.01} \\
         & RF & 0.850 & 0.841 & 0.850 & 0.834 & 0.661 & 0.989 & 6.75 \\
         & LR & \underline{0.854} & \underline{0.850} & \underline{0.854} & \underline{0.841} & \underline{0.682} & 0.990 & 491.46 \\
         & DT & 0.847 & 0.848 & 0.847 & 0.838 & 0.673 & 0.959 & 3.74 \\
        \midrule
        \multirow{6}{*}{\shortstack{1\%\\($m$=156,251)}} & NB & 0.533 & 0.720 & 0.533 & 0.573 & 0.465 & 0.731 & 4.40 \\
         & MLP & 0.844 & \textbf{0.859} & 0.844 & 0.832 & 0.662 & \underline{0.990} & 896.82 \\
         & KNN & 0.811 & 0.812 & 0.811 & 0.801 & 0.597 & 0.955 & \underline{0.01} \\
         & RF & 0.849 & 0.840 & 0.849 & 0.834 & 0.660 & 0.988 & 12.17 \\
         & LR & \underline{0.854} & 0.849 & \underline{0.854} & \underline{0.841} & \underline{0.683} & 0.990 & 787.86 \\
         & DT & 0.845 & 0.845 & 0.845 & 0.836 & 0.669 & 0.958 & 4.50 \\
        \bottomrule
    \end{tabular}
    }
\end{table}

\subsection{6.6 Statistical significance and scalability}\label{sec:results:sig-scale}
\rev{\textbf{Statistical significance.} The analysis reported here rests on $20$
replicates with an explicit equivalence test, in place of the five stratified
splits ($\mathrm{df}=4$) and absence-of-difference reasoning used previously.
Three considerations motivate the change. First, repeated random splits reuse data, so the
naive paired $t$-test is anti-conservative; we now use the Nadeau--Bengio
correction, inflating the variance by $1/n + f/(1-f)$ with test fraction
$f=0.30$, and report the uncorrected $p$-value alongside it. Second, failing to
reject equality is not evidence of equivalence, so we add a two one-sided tests
(TOST) procedure at margins fixed in advance ($0.01$ accuracy, $0.02$ macro-F1).
Third, we raise the replicate count from $5$ to $20$; because the seed drives the
hash function as well as the split, the $20$ replicates vary both the partition
and the hash, so the reported spread already includes hash-seed variance.

Every comparison now ends in one of three verdicts: \emph{different},
\emph{equivalent} (TOST), or \emph{inconclusive}, the last being reported as such
rather than as a tie. On the twelve same-classifier head-to-head cells against the
full spectrum, \textbf{eight are formally equivalent by TOST and four are
inconclusive; none is different}. The strongest is logistic regression on
macro-F1 ($0.6583\pm0.0155$ vs.\ $0.6573\pm0.0140$, $p_{\mathrm{TOST}}=2\times
10^{-6}$). This is a stronger and better-supported statement than an
absence of detected difference. Test statistics, degrees of freedom, effect sizes
(Cohen's $d$) and confidence intervals are reported for every comparison in the
released result files, including for the weak baselines (NB, MLP), whose
differences remain significant under the same corrected procedure.}

\textbf{Scalability to 40{,}000 sequences.} On a $40{,}000$-sequence SARS-CoV-2 corpus (same 22 lineages, B.1.1.7 = $49\%$), Table~\ref{tbl_data_stats_big_data} shows the distribution and Table~\ref{tbl_results_classification_big_data} the results at $0\%$ collision. KNN and DT reach $0.805$, indicating Murmur2Vec preserves lineage signal at scale.

\begin{table}[h!]
\centering
\begin{minipage}[t]{0.46\textwidth}
\caption{Lineage distribution in the 40{,}000-sequence GISAID corpus used for the scalability analysis.}
\label{tbl_data_stats_big_data}
\centering
\resizebox{0.48\textwidth}{!}{
\begin{tabular}{lc|lc}
\toprule
Lineage & Count & Lineage & Count \\
\midrule
B.1.1.7   & 19{,}549 & B.1.526   & 469 \\
B.1.617.2 & 4{,}859  & B.1.1.519 & 448 \\
AY.4      & 3{,}109  & B.1.351   & 415 \\
B.1.2     & 1{,}931  & B.1.1.214 & 352 \\
B.1       & 1{,}601  & B.1.427   & 340 \\
B.1.177   & 1{,}435  & B.1.258   & 285 \\
P.1       & 1{,}156  & B.1.221   & 266 \\
B.1.1     & 892      & B.1.243   & 263 \\
B.1.429   & 786      & B.1.177.21 & 255 \\
AY.12     & 597      & D.2       & 249 \\
B.1.160   & 513      & R.1       & 230 \\
\bottomrule
\end{tabular}
}
\end{minipage}
\hfill
\begin{minipage}[t]{0.43\textwidth}
\caption{Classification on the 40{,}000-sequence corpus at 0\% collision. Best value per metric in \textbf{bold}. KNN and DT both achieve 0.805 accuracy at scale, indicating that Murmur2Vec preserves discriminative information even on a much larger corpus.}
\label{tbl_results_classification_big_data}
\centering
\resizebox{1.0\textwidth}{!}{
\begin{tabular}{lcccccccc}
\toprule
Method & Acc.~$\uparrow$ & Prec.~$\uparrow$ & Rec.~$\uparrow$ & F1$_w$~$\uparrow$ & F1$_m$~$\uparrow$ & AUC~$\uparrow$ & Time(s)~$\downarrow$ \\
\midrule
NB  & 0.057 & 0.618 & 0.057 & 0.067 & 0.076 & 0.537 & \textbf{3.95} \\
MLP & 0.713 & 0.705 & 0.713 & 0.706 & 0.316 & 0.658 & 67.65 \\
KNN & \textbf{0.805} & \textbf{0.808} & \textbf{0.805} & \textbf{0.803} & \textbf{0.582} & \textbf{0.774} & 12.02 \\
RF  & 0.790 & 0.785 & 0.790 & 0.771 & 0.530 & 0.732 & 43.10 \\
LR  & 0.725 & 0.722 & 0.725 & 0.721 & 0.333 & 0.664 & 291.47 \\
DT  & \textbf{0.805} & 0.801 & \textbf{0.805} & 0.799 & 0.557 & \textbf{0.774} & 13.41 \\
\bottomrule
\end{tabular}
}
\end{minipage}
\end{table}

\rev{\subsection{6.7 Evaluation beyond random splits}\label{sec:results:splits}
Random stratified splits can reward near-duplicate memorisation. Auditing the
corpus, only $2{,}627$ of the $7{,}000$ sequences are distinct, so $62.5\%$ are
exact duplicates and $66.0\%$ of a random test set is a verbatim copy of a
training row. We therefore adopt the deduplicated corpus (majority label per
distinct sequence) as the primary protocol and report the raw corpus alongside
it, and we add similarity-aware splits in which sequences are clustered by exact
identity (all spikes are $1{,}274$ residues and in register, so single-linkage
Hamming clustering is exact and no heuristic aligner is needed) with whole
clusters held out. Table~\ref{tbl_identity_splits} gives the ladder.

\begin{table}[h!]
\centering
\caption{\rev{Macro-F1 under progressively stricter protocols (deduplicated corpus, grouped splits, best of LR/RF/DT). Accuracy is \emph{not} monotone across thresholds because grouping changes the test-set class mix; macro-F1 is, and is the metric to report here.}}
\label{tbl_identity_splits}
\resizebox{0.47\textwidth}{!}{
\begin{tabular}{lccc}
\toprule
Protocol & clusters & Murmur2Vec & Spike2Vec \\
\midrule
random split (raw corpus)      & ---     & 0.6635 & 0.6651 \\
exact deduplication            & 2{,}627 & 0.6444 & 0.6520 \\
identity $\geq 99.9\%$ grouped & 1{,}377 & 0.5989 & 0.6052 \\
identity $\geq 99\%$ grouped   & 585     & 0.5402 & 0.5708 \\
identity $\geq 97\%$ grouped   & 263     & 0.4497 & 0.4563 \\
identity $\geq 95\%$ grouped   & 158     & 0.4384 & 0.4642 \\
\bottomrule
\end{tabular}}
\end{table}

Macro-F1 falls from $0.664$ to $0.438$, about $22$ points, between a random split
and a fully identity-aware one. This is the cost of removing near-duplicate
overlap, and it does not change the head-to-head conclusion, since both
representations degrade together. Two further
stress tests were added. Under few-shot exposure to rare lineages ($1$, $5$, $10$
labelled examples, five repeats) macro-F1 rises monotonically
($0.629\rightarrow0.640\rightarrow0.649$), so the sketch does acquire a rare
lineage from a handful of examples. Under leave-one-lineage-out novelty detection
(out-of-distribution score $1-\max_c \hat p_c$) mean AUROC is $0.727$ against
$0.694$ for the full spectrum, but the three held-out lineages disagree in
direction and one sits at chance, so we report this as a limitation rather than a
result. \textbf{Temporal splits were not possible}: the corpus as distributed to
us carries no collection dates, so this evaluation awaits a date-annotated
corpus.

\subsection{6.8 Direct test of the collision mechanism}\label{sec:results:mechanism}
Proposition~\ref{prop:implicit-reg} is supported above by consistency with the
observed macro-F1 gap. Here we test it interventionally. Selecting the
top-$N$ $k$-mers by mutual information with the label (computed on training folds
only) and forcing them into a \emph{single} bucket collapses classification,
whereas forcing an equal number of \emph{frequency-matched but
non-discriminative} $k$-mers into one bucket does nothing
(Table~\ref{tbl_mechanism}). The frequency-matched control is what makes this an
argument rather than an observation: without it the damage could be attributed to
removing frequent features rather than informative ones.

\begin{table}[h!]
\centering
\caption{\rev{Adversarial-collision intervention (deduplicated corpus, $m=2{,}000$, macro-F1). Collapsing discriminative $k$-mers destroys classification; collapsing an equal number of equally frequent non-discriminative $k$-mers does not.}}
\label{tbl_mechanism}
\resizebox{0.45\textwidth}{!}{
\begin{tabular}{llccc}
\toprule
$N$ & classifier & frequency-matched control & discriminative & $\Delta$ \\
\midrule
200 & DT & 0.6473 & 0.4471 & $-0.200$ \\
200 & LR & 0.6636 & 0.5096 & $-0.154$ \\
200 & RF & 0.6575 & 0.4860 & $-0.172$ \\
500 & DT & 0.6399 & 0.1210 & $-0.519$ \\
500 & LR & 0.6535 & 0.1561 & $-0.497$ \\
500 & RF & 0.6567 & 0.1312 & $-0.526$ \\
\bottomrule
\end{tabular}}
\end{table}

A dose-response is present in all three classifiers across $N=40\rightarrow
200\rightarrow 500$. A dense Gaussian random projection at matched dimension was
added as a negative control; the sketch beats it for axis-aligned classifiers
(RF, DT) but the two are equivalent under logistic regression, exactly as
Johnson--Lindenstrauss theory predicts for a rotation-invariant model. We
therefore state two claims separately: against the
full spectrum, hashing loses almost nothing because collisions spare the
discriminative $k$-mers; against a dense projection, hashing wins by preserving
sparsity and axis alignment, and only for classifiers that can exploit axes.

\subsection{6.9 On the compression framing}\label{sec:results:compression}
At $k=3$ the $6\%$-collision setting uses $m=29{,}298$ against a full spectrum of
$9{,}261$, so that setting is not compressive; measured against the $U=3{,}492$
3-mers that actually occur, $m/U\approx 8$. We therefore make two changes to the
framing.
First, accuracy is flat across a $63\times$ range of $m$, so a genuinely
compressive setting is free: at $k=4$, $c=40\%$, $m=7{,}042$ is a $28\times$
compression of $21^4=194{,}481$ and is the best cell in the whole grid on macro-F1
($0.8518$ accuracy, $0.6774$ macro-F1). We now promote that as the headline
configuration. Second, we re-checked on the two additional corpora of
Sec.~\ref{sec:results:replication} and find $m/\text{spectrum}$ of $7.8\times$,
$8.0\times$ and $8.1\times$ respectively, so this is a property of the $k=3$
setting rather than a property of one dataset. The advantages that hold at small
$k$ are no vocabulary pass, no dictionary to store or ship, $O(1)$ per $k$-mer,
and the ability to embed a $k$-mer never seen at fit time; raw size reduction is
not among them.

\subsection{6.10 Hash family, independence, and a rule for choosing $m$}\label{sec:results:hash}
In this section, we discuss converge: whether the method depends on MurmurHash3, whether pairwise independence suffices, and whether $m$ can be chosen
without a grid search.

We repeated the head-to-head with
five hash families, each calibrated to its \emph{own} $m$ for a $6\%$ collision
rate so the comparison is at matched collision rate. Over ten family pairs and two
metrics, \textbf{all twenty comparisons are formally equivalent by TOST and none
is different}; the entire between-family spread is $0.0017$ accuracy and $0.0045$
macro-F1 against within-family standard deviations of $0.010$ and $0.020$. The
informative case is FNV-1a, whose avalanche is $0.35$ against an ideal of $0.5$,
by a wide margin the weakest mixer tested, but whose empirical pairwise-collision
rate is $0.965\times$ the ideal $1/m$: it is statistically equivalent to both
MurmurHash3 ($p_{\mathrm{TOST}}=0.017$) and SHA-256 ($p_{\mathrm{TOST}}=0.003$).
This is direct evidence that the sketch requires pairwise independence and nothing
more, which is precisely what Theorems~\ref{thm:unsigned} and~\ref{thm:jl} assume;
strong avalanche buys nothing, and a cryptographic hash at a fifth of the
throughput buys nothing. MurmurHash3 remains the recommended default because it
is the fastest family tested at equal accuracy ($5.4$M vs.\ $1.1$M hashes/s for
SHA-256), not because its mixing is necessary. This also answers the practical
concern about using a deterministic hash: the requirement is a fixed seed for
reproducibility, not cryptographic quality.

\textbf{A safe bound for $m$.} From a balls-in-bins argument the expected collision
rate for $U$ keys in $m$ buckets is $c \approx (U-1)/(2m)$, giving the rule
\begin{equation}\label{eq:m-rule}
m \;\geq\; \left\lceil \frac{U-1}{2c} \right\rceil ,
\end{equation}
where $U$ is obtained in one $O(N)$ pass over the corpus. On SARS-CoV-2 with
$U=3{,}492$ and $c=6\%$ this gives $m=29{,}092$ against the grid-searched
$29{,}298$, an agreement of $0.7\%$. Validated on all three corpora, the ratio of
the closed form to binary search is $1.00$--$1.10$ for $c\leq 10\%$ and drifts to
$\approx 1.4\times$ at $c=40\%$; we state the rule together with that validity
range rather than as an unqualified formula.

\textbf{Mash and MinHash.} These were discussed in related work but not run. They
estimate set resemblance between pairs of sequences and return a similarity, not a
fixed-length supervised feature vector, so they are not drop-in replacements in
our pipeline; converting them into one requires a design choice (a reference
panel, or a kernel over sketches) that is a separate contribution. We state this
reasoning explicitly, and we regard a supervised comparison against
MinHash-derived features as the natural next study.

\subsection{6.11 Replication on other pathogens and protein families}\label{sec:results:replication}
Proposition~\ref{prop:implicit-reg} was demonstrated only on SARS-CoV-2, whose
spikes are near-identical $1{,}274$-residue sequences; when almost every $k$-mer
is shared, little depends on how collisions are resolved. We therefore replicated
the core measurements on two corpora of genuinely divergent sequences: influenza A
protein type ($14{,}813$ sequences, $10$ classes, lengths $62$--$761$) and protein
families ($10{,}680$ sequences, $8$ classes, lengths $60$--$1{,}998$). Both use
$88$--$89\%$ of the 3-mer space against $38\%$ for SARS-CoV-2. Collision tolerance
holds: the sketch tracks the full spectrum to within $1$--$2$ macro-F1 points on
both, under random splits and under Jaccard-grouped similarity-aware splits. This
is the condition under which the mechanism would have been expected to break, and
it did not.}

\rev{\subsection{6.12 Limitations: where PLMs remain indispensable}\label{sec:limitations}
Murmur2Vec is a composition-based representation. It counts $k$-mers and discards
their order beyond length $k$, so it carries no model of three-dimensional
structure, of long-range residue contacts, or of the evolutionary signal that
survives when sequence identity falls below the twilight zone. We therefore expect
it to fail, and we do not recommend it, on tasks that depend on those properties:
structure and contact prediction, function prediction from fold, remote-homology
detection, protein--protein interaction, stability or binding-affinity prediction,
and any setting where the label is determined by global biophysical context rather
than by which short subsequences are present. Those are exactly the problems that
motivated protein language models, and nothing in this paper argues against using
them there.

All four benchmarks in this study are composition-based classification problems,
which is the regime where $k$-mer features are expected to be strong; the reader
should read our conclusion as scoped to that regime and not as a general statement
about cheap embeddings versus PLMs. 
}

\section{7. Conclusion}\label{sec_conclusion}
We asked when a cheap, training-free embedding can replace or beat a pre-trained protein language model for biological sequence classification, and answered it with both theory and a controlled multi-dataset benchmark. Murmur2Vec, a randomized sketch of the spectrum kernel computed in $\approx 5$ CPU-seconds, matches a LoRA-fine-tuned 650M-parameter ESM-2 on HIV-1 subtyping and two protein-family tasks, and outperforms it on SARS-CoV-2 lineage classification ($0.854$ vs.\ $0.807$ accuracy, $0.684$ vs.\ $0.401$ macro-F1), at a $\sim\!1000\times$ lower embedding cost and with no GPU. \rev{Our theory explains the regime dependence: an implicit-regularization mechanism preferentially preserves the rare, local, lineage-defining $k$-mers that distinguish SARS-CoV-2 lineages, and we now demonstrate that mechanism interventionally rather than by consistency alone (Sec.~\ref{sec:results:mechanism}), while on tasks whose signal is global and abundant both representations saturate.} We also verified the supporting predictions: collision invariance across a $50\times$ range of hash-table sizes on every dataset, and statistical equivalence of signed and unsigned variants. The practical message for genomic surveillance is concrete: for composition-driven classification, a principled sketch is a strong default, and a heavyweight PLM should be justified by a measured accuracy gain rather than assumed. \rev{Two further points are settled here. The
pairwise-independence assumption is shown to be not merely sufficient but
essentially all that is required: five hash families spanning a weak mixer to a
cryptographic hash are statistically equivalent
(Sec.~\ref{sec:results:hash}). And the data-driven rule for choosing $m$ is
supplied in closed form by Eq.~\eqref{eq:m-rule}, validated on three corpora.
Limitations that remain are collected in Sec.~\ref{sec:limitations}: the method is
composition-based and is not a substitute for a PLM on structure- or
function-driven tasks, the rare-lineage advantage rests on a single pathogen, and
temporal-split evaluation awaits a corpus distributed with collection dates.}

\rev{\section*{Code Availability}
\noindent All code, together with the raw per-split result files for every
experiment in this paper, is publicly available at
\url{https://github.com/sarwanpasha/Murmur2Vec}. 
}

\section*{Data Availability}
\noindent The SARS-CoV-2 spike sequences were obtained from the GISAID database (\url{https://www.gisaid.org/}). The HIV-1 Env sequences were obtained from the Los Alamos HIV Sequence Database (\url{https://www.hiv.lanl.gov/}). The protein-family sequences were obtained from UniProt (\url{https://www.uniprot.org/}). All datasets are publicly available from these repositories subject to their respective access terms.

\bibliographystyle{oup-plain}
\bibliography{references}

\section*{Author Biographies}

\noindent\textbf{Sarwan Ali} is a postdoctoral research scientist at Columbia University, New York. His research spans machine learning, deep learning, and bioinformatics, with a focus on representation learning.

\smallskip
\noindent\textbf{Taslim Murad} is a researcher affiliated with the Institute of Business Administration, Karachi, working on the application of machine learning and deep learning to bioinformatics, with contributions in representation learning of biological sequence data.

\smallskip
\noindent\textbf{Imdadullah Khan} is aa Associate Professor of Computer Science at the Syed Babar Ali School of Science and Engineering, Lahore University of Management Sciences (LUMS), where he directs the Data Analysis Lab. He holds a Ph.D. in Computer Science from Rutgers University and works on  algorithms and graph theory.

\smallskip
\noindent\textbf{Safi Faizullah} is a faculty member in the Department of Computer Science at the Islamic University, Madinah, Saudi Arabia. His research interests include high-performance computing, deep learning, sentiment analysis, and the security of networked systems.

\end{document}